\documentclass[final]{cvpr}

\usepackage{caption}
\usepackage{times}
\usepackage{epsfig}
\usepackage{graphicx}
\usepackage{amsmath}
\usepackage{amssymb}
\usepackage{comment}
\usepackage{cuted}
\usepackage{capt-of}
\usepackage[font={small}]{caption}
\usepackage{afterpage}
\usepackage{amsmath}
\usepackage{enumitem}
\usepackage{booktabs}
\usepackage{siunitx}

\usepackage[pagebackref=true,breaklinks=true,letterpaper=true,colorlinks,bookmarks=false]{hyperref}

\makeatletter
\newcommand*{\transpose}{%
	{\mathpalette\@transpose{}}%
}
\newcommand*{\@transpose}[2]{%
	\raisebox{\depth}{$\m@th#1\intercal$}%
}
\makeatother

\DeclareMathOperator*{\argmin}{arg\,min}

\let\depth\relax
\begin{document}

\title{LOHO: Latent Optimization of Hairstyles via Orthogonalization}

\author{Rohit Saha\textsuperscript{1,2}\thanks{Corresponding Author: \texttt{rohitsaha@cs.toronto.edu}}
	\and
	\hspace{-0.1in}Brendan Duke\textsuperscript{1,2}
	\and
	\hspace{-0.1in}Florian Shkurti\textsuperscript{1,4}
	\and
	\hspace{-0.1in}Graham W.~Taylor\textsuperscript{3,4}
	\and
	\hspace{-0.1in}Parham Aarabi\textsuperscript{1,2}
	\and
	\textsuperscript{1}University of Toronto\qquad
	\textsuperscript{2}Modiface, Inc.\qquad
	\textsuperscript{3}University of Guelph\qquad
	\textsuperscript{4}Vector Institute
}

\maketitle
\begin{strip}\centering
	\vspace{-0.5in}
	\includegraphics[width=\textwidth]{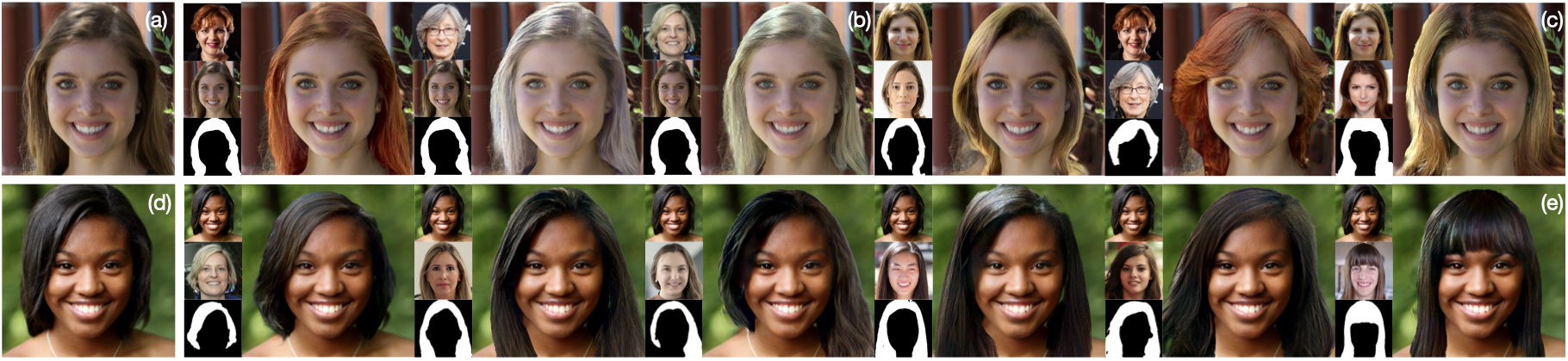}
	\vspace{-.3in}
	\captionof{figure}{\textbf{Hairstyle transfer samples synthesized using LOHO.}
		For given portrait images (a) and (d), LOHO is capable of manipulating hair attributes based on multiple input conditions.
		Inset images represent the target hair attributes in the order: appearance and style, structure, and shape.
		LOHO can transfer appearance and style (b), and perceptual structure (e) while keeping the background unchanged.
		Additionally, LOHO can change multiple hair attributes simultaneously and independently (c).}
	\label{fig:mix_67172}
\end{strip}

\begin{abstract}
	Hairstyle transfer is challenging due to hair structure differences in the source and target hair.
	Therefore, we propose Latent Optimization of Hairstyles via Orthogonalization (LOHO), an optimization-based approach using GAN inversion to infill missing hair structure details in latent space during hairstyle transfer.
	Our approach decomposes hair into three attributes: perceptual structure, appearance, and style, and includes tailored losses to model each of these attributes independently.
	Furthermore, we propose two-stage optimization and gradient orthogonalization to enable disentangled latent space optimization of our hair attributes.
	Using LOHO for latent space manipulation, users can synthesize novel photorealistic images by manipulating hair attributes either individually or jointly, transferring the desired attributes from reference hairstyles.
	LOHO achieves a superior FID compared with the current state-of-the-art (SOTA) for hairstyle transfer.
	Additionally, LOHO preserves the subject's identity comparably well according to PSNR and SSIM when compared to SOTA image embedding pipelines.
	Code is available at~\url{https://github.com/dukebw/LOHO}.
\end{abstract}

\section{Introduction}

We set out to enable users to make semantic and structural edits
to their portrait image with fine-grained control.
As a particular challenging and commercially appealing example, we study hairstyle transfer,
wherein a user can transfer hair attributes from multiple independent source
images to manipulate their own portrait image.
Our solution, Latent Optimization of Hairstyles via Orthogonalization (LOHO), is a two-stage optimization process in the latent space of
a generative model, such as a generative adversarial network
(GAN)~\cite{goodfellow2014gan, karras2017progressive}.
Our key technical contribution is that we control attribute transfer by orthogonalizing the gradients of our
transferred attributes so that the application of one attribute does not
interfere with the others.

Our work is motivated by recent progress in GANs, enabling both
conditional~\cite{isola2017pix2pix, wang2018pix2pixhd} and
unconditional~\cite{karras2019astylebased} synthesis of photorealistic images.
In parallel, recent works have achieved impressive latent space manipulation by
learning disentangled feature representations~\cite{pidhorskyi2020adversarial},
enabling photorealistic global and local image manipulation.
However, achieving controlled manipulation of attributes of the synthesized
images while maintaining photorealism remains an open challenge.

Previous work on hairstyle transfer~\cite{tan2020michigan} produced realistic
transfer of hair appearance using a complex pipeline of GAN generators, each
specialized for a specific task such as hair synthesis or background
inpainting.
However, the use of pretrained inpainting networks to fill holes left over by
misaligned hair masks results in blurry artifacts.
To produce more realistic synthesis from transferred hair shape, we can infill missing shape and structure details by invoking the prior
distribution of a single GAN pretrained to generate faces.

To achieve photorealistic hairstyle transfer even under said source-target hair
misalignment we propose Latent Optimization of Hairstyles via Orthogonalization
(LOHO).
LOHO directly optimizes the extended latent space and the noise space of a pretrained StyleGANv2~\cite{karras2020stylegan2}.
Using carefully designed loss functions, our approach decomposes hair into three attributes: perceptual structure, appearance, and  style.
Each of our attributes is then modeled individually, thereby allowing better control over the synthesis process.
Additionally, LOHO significantly improves the quality of synthesized images by employing two-stage optimization, where each
stage optimizes a subset of losses in our objective function.
Our key insight is that some of the losses, due to their similar design, can only be optimized sequentially and not jointly.
Finally, LOHO uses gradient orthogonalization to explicitly disentangle hair attributes during the optimization process.

Our main contributions are:
\begin{itemize}
	\item We propose a novel approach to perform hairstyle transfer by optimizing StyleGANv2's extended latent space and noise space.
	\item We propose an objective that includes multiple losses catered to model each key hairstyle attribute. %
	\item We propose a two-stage optimization strategy %
	      that leads to significant improvements in the photorealism of synthesized images.
	\item We introduce gradient orthogonalization, a general method to jointly optimize attributes in latent space without interference.
	      We demonstrate the effectiveness of gradient orthogonalization both qualitatively and quantitatively.
	\item We apply our novel approach to perform hairstyle transfer on in-the-wild portrait images and compute the Fréchet Inception Distance (FID) score.
	      FID is used to evaluate generative models by calculating the distance between Inception~\cite{szegedy2016inceptionv3} features
	      for real and synthesized images in the same domain.
	      The computed FID score shows that our approach outperforms the current state-of-the-art (SOTA) hairstyle transfer results.
\end{itemize}

\section{Related Work}

\textbf{Generative Adversarial Networks.} Generative models, in particular GANs, have been very successful across various computer vision applications such as image-to-image translation~\cite{isola2017pix2pix, wang2018pix2pixhd, zhu2017cycle}, video generation~\cite{wang2018vid2vid, wang2018fewshotvid2vid, chan2019dance}, and data augmentation for discriminative tasks such as object detection~\cite{li2017objectdetection}.
GANs~\cite{karras2017progressive,brock2018large} transform a latent code to an image by learning the underlying distribution of training data.
A more recent architecture, StyleGANv2~\cite{karras2020stylegan2}, has set the benchmark for generation of photorealistic human faces.
However, training such networks requires significant amounts of data, significantly increasing the barrier to train SOTA GANs for specific use cases such as hairstyle transfer.
Consequently, methods built using pretrained generators are becoming the de facto standard for executing various image manipulation tasks.
In our work, we leverage~\cite{karras2020stylegan2} as an expressive pretrained face synthesis
model, and outline our optimization approach for using pretrained generators
for controlled attribute manipulation.

\textbf{Latent Space Embedding.} Understanding and manipulating the latent space of GANs via inversion has become an active field of research. %
GAN inversion involves embedding an image into the latent space of a GAN such that the synthesized image resulting from that latent embedding is the most accurate reconstruction of the original image.
I2S~\cite{abdal2019image2stylegan} is a framework able to reconstruct an image by optimizing the extended latent space $\mathcal{W}^+$ of a pretrained StyleGAN~\cite{karras2019astylebased}.
Embeddings sampled from $\mathcal{W}^+$ are the concatenation of 18 different 512-dimensional \textit{w} vectors, one for each layer of the StyleGAN architecture.
I2S++~\cite{abdal2020i2s++} further improved the image reconstruction quality by additionally optimizing the noise space $\mathcal{N}$.
Furthermore, including semantic masks in the I2S++ framework allows users to perform tasks such as image inpainting and global editing.
Recent methods~\cite{guan2020collaborative, richardson2020encoding, zhu2020indomain} learn an encoder to map inputs from the image space directly to $\mathcal{W}^+$ latent space.
Our work follows GAN inversion, in that our method optimizes the more recent StyleGANv2's $\mathcal{W}^+$ space and noise space $\mathcal{N}$ to perform semantic editing of hair on portrait images.
We further propose a GAN inversion algorithm for simultaneous manipulation of
spatially local attributes, such as hair structure, from multiple sources while
preventing interference between the attributes' different competing objectives.

\begin{figure}[t]
	\begin{center}
		\includegraphics[width=1.0\linewidth]{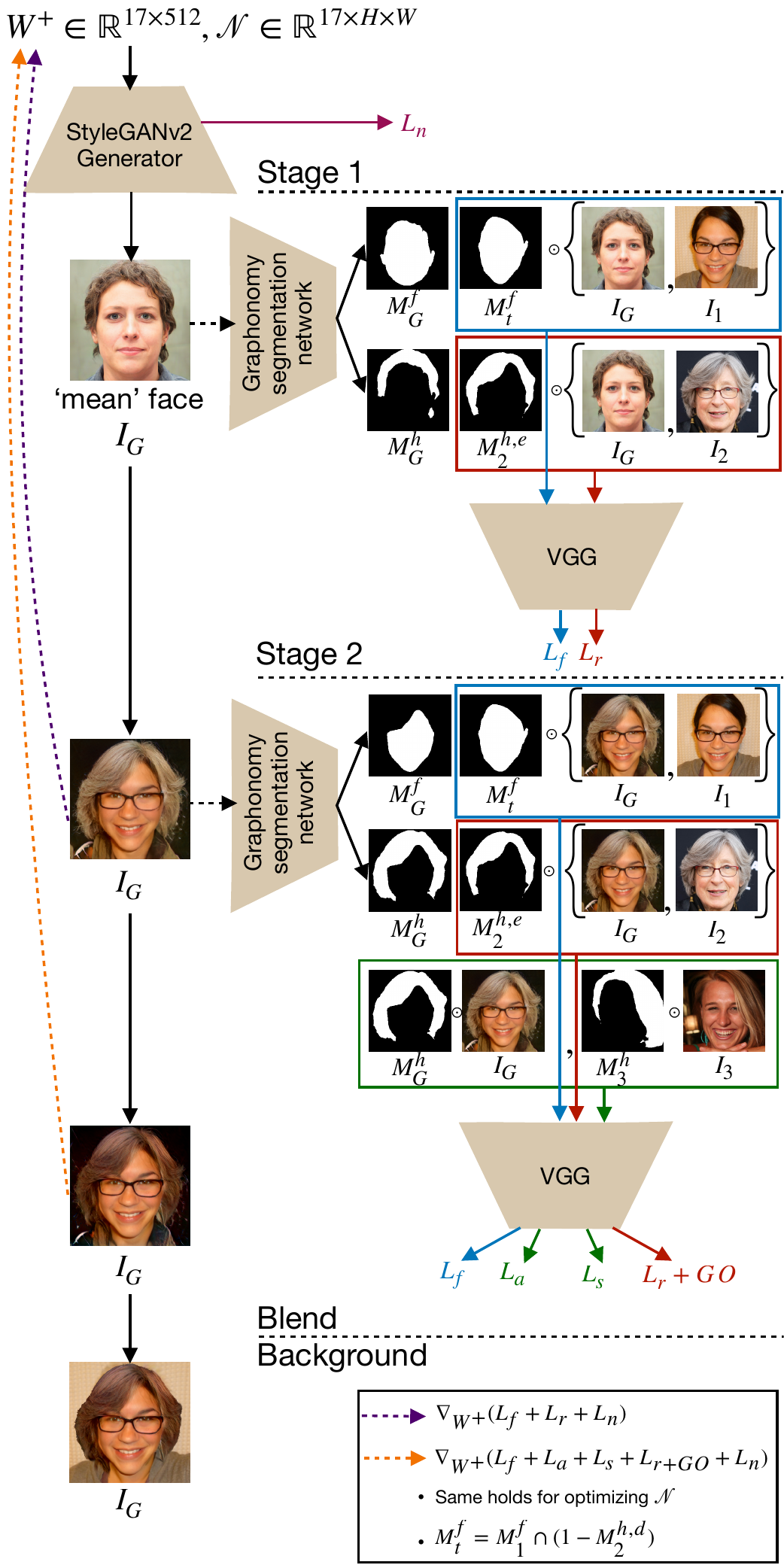}
		\vspace{-.4in}
	\end{center}
	\caption{\textbf{LOHO}. Starting with the 'mean' face, LOHO reconstructs the target identity and the target perceptual structure of hair in Stage 1.
		In Stage 2, LOHO transfers the target hair style and appearance, while maintaining the perceptual structure via Gradient Orthogonalization (GO).
		Finally, $I_G$ is blended with $I_1$'s background.
		(Figure best viewed in colour)
	}
	\label{fig:overview}
	\vspace{-0.2in}
\end{figure}

\textbf{Hairstyle Transfer.} Hair is a challenging part of the human face to model and synthesize.
Previous work on modeling hair includes capturing hair geometry~\cite{chai2012transferring, chai2013animation, chai2015lighting, weng2013morphing}, and using this hair geometry downstream for interactive hair editing.
However, these methods are unable to capture key visual factors, thereby compromising the result quality.
Although recent work~\cite{jo2019scfegan,lee2020maskgan,kim2018realtimehair} showed progress on using GANs for hair generation, these methods do not allow for intuitive control over the synthesized hair.
MichiGAN~\cite{tan2020michigan} proposed a conditional synthesis GAN that allows controlled manipulation of hair.
MichiGAN disentangles hair into four attributes by designing deliberate mechanisms and representations and produces SOTA results for hair appearance change.
Nonetheless, MichiGAN has difficulty handling hair transfer scenarios with
arbitrary shape change.
This is because MichiGAN implements shape change using a separately
trained inpainting network to fill ``holes'' created during the hair
transfer process.
In contrast, our method invokes the prior distribution of a pretrained GAN to
``infill'' in latent space rather than pixel space.
As compared to MichiGAN, our method produces more
realistic synthesized images in the challenging case where hair shape changes.

\section{Methodology}

\subsection{Background}

We begin by observing the objective function proposed in Image2StyleGAN++ (I2S++)~\cite{abdal2020i2s++}:
\begin{equation}
	\begin{split}
		L &= \lambda_s L_{\text{style}}(M_s, G(w, n), y)  \\
		&\quad + \lambda_p L_{\text{percept}}(M_p, G(w, n), x) \\
		&\quad + \frac{\lambda_{\text{mse}1}}{N}{\lVert M_m  \odot (G(w,n)-x)\rVert}_2^{2} \\
		&\quad + \frac{\lambda_{\text{mse}2}}{N}{\lVert (1 - M_m) \odot (G(w,n)-y)\rVert}_2^{2}
	\end{split}
	\label{eq:abdal}
\end{equation}

\noindent where $w$ is an embedding in the extended latent space $\mathcal{W}^+$ of StyleGAN, $n$ is a noise vector embedding, $M_s$, $M_m$, and $M_p$ are binary masks to specify image regions contributing to the respective losses, $\odot$ denotes the Hadamard product, $G$ is the StyleGAN generator, $x$ is the image that we want to reconstruct in mask $M_m$, and $y$ is the image that we want to reconstruct outside $M_m$, i.e., in $(1 - M_m$).

~\cite{abdal2020i2s++} use variations on the I2S++ objective function in Equation~\ref{eq:abdal} to improve image reconstruction, image crossover, image inpainting, local style transfer, and other tasks.
In our case for hairstyle transfer we want to do both image crossover and image inpainting.
Transferring one hairstyle to another person requires crossover, and the leftover region where the original person's hair used to be requires inpainting.

\subsection{Framework}

For the hairstyle transfer problem, suppose we have three portrait images of people: $I_1$, $I_2$ and $I_3$.
Consider transferring person 2's ($I_2$'s) hair shape and structure, and person 3's ($I_3$'s) hair appearance and style to person 1 ($I_1$).
Let $M^f_1$ be ${I_1}$'s binary face mask, and $M^h_1$, $M^h_2$ and $M^h_3$ be ${I_1}$'s, ${I_2}$'s, and ${I_3}$'s binary hair masks.
Next, we separately dilate and erode $M^h_2$ by roughly 20\% to produce the dilated version, $M^{h,d}_2$, and the eroded version, $M^{h,e}_2$.
Let $M^{h,\text{ir}}_2 \equiv M^{h,d}_2 - M^{h,e}_2$ be the ignore region that requires inpainting.
We do not optimize $M^{h,\text{ir}}_2$, and rather invoke StyleGANv2 to inpaint relevant details in this region.
This feature allows our method to perform hair shape transfer in situations where person 1 and person 2's hair shapes are misaligned.

In our method the background of $I_1$ is not optimized.
Therefore, to recover the background, we soft-blend $I_1$'s background with the synthesized image's foreground (hair and face).
Specifically, we use GatedConv~\cite{yu2019gatedconv} to inpaint the masked out foreground region of $I_1$, following which we perform the blending (Figure~\ref{fig:overview}).

\subsection{Objective}

To perform hairstyle transfer, we define the losses that we use to supervise relevant regions of the synthesized image.
To keep our notation simple, let $I_G \equiv G(\mathcal{W}^+,\mathcal{N})$ be the synthesized image, and $M^f_G$ and $M^h_G$ be its corresponding face and hair regions.

\textbf{Identity Reconstruction.} In order to reconstruct person 1's identity we use the Learned Perceptual Image Patch Similarity (LPIPS)~\cite{zhang2018lpips} loss.
LPIPS is a perceptual loss based on human similarity judgements and, therefore, is well suited for facial reconstruction.
To compute the loss, we use pretrained VGG~\cite{simonyan15vgg} to extract high-level features~\cite{johnson2016perceptual} for both $I_1$ and $I_G$.
We extract and sum features from all 5 blocks of VGG to form our face reconstruction objective
\begin{multline}
	L_{f} =  \frac{1}{5} \sum_{b=1}^{5} \textsc{Lpips} \bigl[ \textrm{VGG}^{b} (I_1 \odot (M^f_1 \cap (1 - M^{h,d}_2))),\\
	\textrm{VGG}^{b} (I_G \odot (M^f_1 \cap (1 - M^{h,d}_2))) \bigr]
	\label{eq:lpips-face}
\end{multline}

\noindent where $b$ denotes a VGG block, and $M^f_1 \cap (1 - M^{h,d}_2)$ represents the target mask, calculated as the overlap between $M^f_1$, and the foreground region of the dilated mask $M^{h,d}_2$.
This formulation places a soft constraint on the target mask.

\textbf{Hair Shape and Structure Reconstruction.} To recover person 2's hair information, we enforce supervision via the LPIPS loss.
However, na\"ively using $M^h_2$ as the target hair mask can cause the generator to synthesize hair on undesirable regions of $I_G$.
This is especially true when the target face and hair regions do not align well.
To fix this problem, we use the eroded mask, $M^{h,e}_2$, as it places a soft constraint on the target placement of synthesized hair.
$M^{h,e}_2$, combined with $M^{h,\text{ir}}_2$, allow the generator to handle misaligned pairs by inpainting relevant information in non-overlapping regions.
To compute the loss, we extract features from blocks 4 and 5 of VGG corresponding to hair regions of $I_2$, $I_G$ to form our hair perceptual structure objective
\begin{equation}
	\begin{split}
		L_{r} = \frac{1}{2} \sum_{b \in \{4, 5\}}  \textsc{Lpips} \bigl[ & \textrm{VGG}^{b} (I_2 \odot M^{h,e}_2),\\
			& \textrm{VGG}^{b} (I_G \odot M^{h,e}_2) ]
		\label{eq:lpips-hairrec}
	\end{split}
\end{equation}

\textbf{Hair Appearance Transfer.} Hair appearance refers to the globally consistent colour of hair that is independent of hair shape and structure.
As a result, it can be transferred from samples of different hair shapes.
To transfer the target appearance, we extract 64 feature maps from the shallowest layer of VGG ($relu1\_1$) as it best accounts for colour information.
Then, we perform average-pooling within the hair region of each feature map to discard spatial information and capture global appearance.
We obtain an estimate of the mean appearance $A$ in $\mathbb{R}^{64 \times 1}$ as $A(x, y) = \sum \frac{\phi(x) \odot y}{|y|}$, where $\phi(x)$ represents the 64 VGG feature maps of image x, and y indicates the relevant hair mask.
Finally, we compute the squared $L_2$ distance to give our hair appearance objective
\begin{gather}
	\label{eq:appearance}
	L_{a} = {\lVert A(I_3, M^h_3) - A(I_G, M^h_G)\rVert}_2^{2}
\end{gather}

\textbf{Hair Style Transfer.} In addition to the overall colour, hair also contains finer details such as wisp styles, and shading variations between hair strands.
Such details cannot be captured solely by the appearance loss that estimates the overall mean.
Better approximations are thus required to compute the varying styles between hair strands.
The Gram matrix~\cite{gatys2016style} captures finer hair details by calculating the second-order associations between high-level feature maps.
We compute the Gram matrix after extracting features from layers: \{$relu1\_2, relu2\_2, relu3\_3, relu4\_3$\} of VGG
\begin{equation}
	\mathcal{G}^l(\gamma^l) =  {\gamma^l}^{\transpose} \gamma^l
	\label{eq:style1}
\end{equation}
\noindent where, $\gamma^l$ represents feature maps in $\mathbb{R}^{HW \times C}$ that are extracted from layer $l$, and~$\mathcal{G}^l$ is the Gram matrix at layer~$l$.
Here, $C$ represents the number of channels, and $H$ and $W$ are the height and width.
Finally, we compute the squared $L_2$ distance as
\begin{multline}
	L_{s} =  \frac{1}{4} \sum_{l=1}^{4} \lVert \mathcal{G}^l(\textrm{VGG}^l(I_3 \odot M^h_3))\\
	- \mathcal{G}^l(\textrm{VGG}^l(I_G \odot M^h_G)) \rVert_2^2
	\label{eq:style2}
\end{multline}

\textbf{Noise Map Regularization.} Explicitly optimizing the noise maps $n \in \mathcal{N}$ can cause the optimization to inject actual signal into them.
To prevent this, we introduce regularization terms of noise maps~\cite{karras2020stylegan2}.
For each noise map greater than $8 \times 8$, we use a pyramid down network to reduce the resolution to $8 \times 8$.
The pyramid network averages $2 \times 2$ pixel neighbourhoods at each step.
Additionally, we normalize the noise maps to be zero mean and unit variance, producing our noise objective
\begin{equation}
	\begin{split}
		L_{n} = & \sum_{i,j} \left[ \frac{1}{r_{i,j}^2} \cdot \sum_{x,y} n_{i,j}(x,y) \cdot n_{i,j}(x-1,y) \right]^2 \\
		+ & \sum_{i,j} \left[ \frac{1}{r_{i,j}^2} \cdot \sum_{x,y} n_{i,j}(x,y) \cdot n_{i,j}(x,y-1) \right]^2
	\end{split}
	\label{eq:lpips-hairrec}
\end{equation}

\noindent where $n_{i,0}$ represents the original noise map and $n_{i,j>0}$ represents the downsampled versions. Similarly, $r_{i,j}$ represents the resolution of the original or downsampled noise map.

Combining all the losses the overall optimization objective is
\begin{equation}
	\begin{split}
		L = \underset{\{\mathcal{W}^+,\mathcal{N}\}}{\argmin} \bigl[\lambda_f L_{f} & + \lambda_r L_{r} + \lambda_a L_{a} \\
			& + \lambda_s L_{s} + \lambda_n L_{n}\bigr]
	\end{split}
	\label{eq:overall}
\end{equation}

\begin{figure}[!t]
	\begin{center}
		\includegraphics[width=1.0\linewidth]{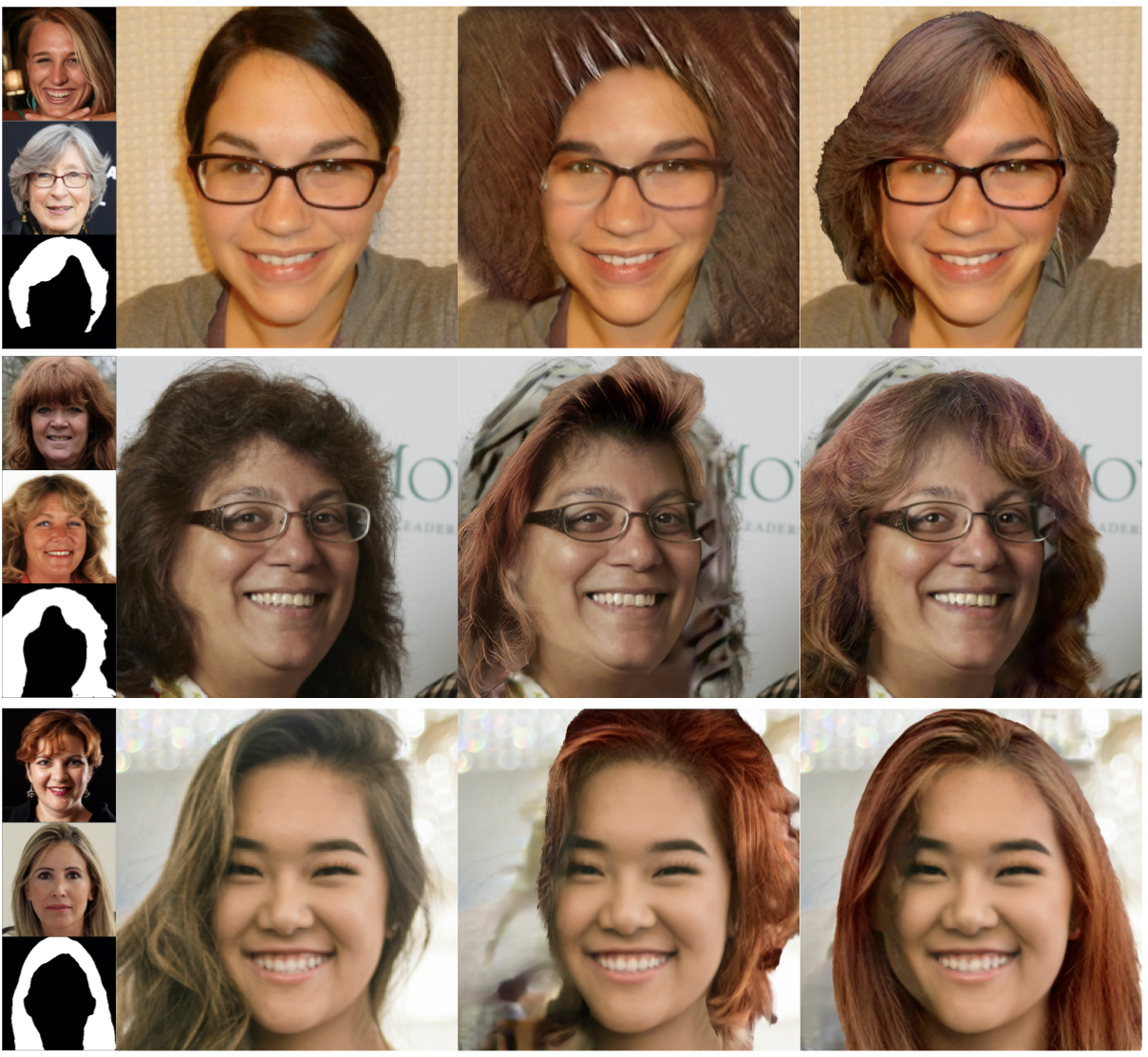}
		\vspace{-.4in}
	\end{center}
	\caption{\textbf{Effect of two-stage optimization.}
		\textbf{Col 1 (narrow)}: Reference images.
		\textbf{Col 2}: Identity person.
		\textbf{Col 3}: Synthesized image when losses arre optimized jointly.
		\textbf{Col 4}: Image synthesized via two-stage optimization + gradient orthogonalization.}
	\label{fig:1-2-stage}
	\vspace{-0.15in}
\end{figure}

\subsection{Optimization Strategy} \label{optimGO}

\textbf{Two-Stage Optimization.} Given the similar nature of the losses $L_{r}$, $L_{a}$, and $L_{s}$, we posit that jointly optimizing all losses from the start will cause person 2's hair information to compete with person 3's hair information, leading to undesirable synthesis.
To mitigate this issue, we optimize our overall objective in two stages.
In stage~$1$, we reconstruct only the target identity and hair perceptual structure, i.e., we set $\lambda_a$ and $\lambda_s$ in Equation~\ref{eq:overall} to zero.
In stage~$2$, we optimize all the losses except $L_r$; stage~$1$ will provide a better initialization for stage~$2$, thereby leading the model to convergence.

However, this technique in itself has a drawback.
There is no supervision to maintain the reconstructed hair perceptual structure after stage 1.
This lack of supervision allows StyleGANv2 to invoke its prior distribution to inpaint or remove hair pixels, thereby undoing the perceptual structure initialization found in stage~$1$.
Hence, it is necessary to include $L_{r}$ in stage~$2$ of optimization.

\textbf{Gradient Orthogonalization.} $L_{r}$, by design, captures all hair attributes of person 2: perceptual structure, appearance, and style.
As a result, $L_{r}$'s gradient competes with the gradients corresponding to the appearance and style of person 3.
We fix this problem by manipulating $L_{r}$'s gradient such that its appearance and style information are removed.
More specifically, we project $L_{r}$'s perceptual structure gradients onto the vector subspace orthogonal to its appearance and style gradients.
This allows person 3's hair appearance and style to be transferred while preserving person 2's hair structure and shape.

Assuming we are optimizing the $\mathcal{W}^+$ latent space, the gradients computed are
\begin{equation}
	g_{R_2} = \nabla_{\mathcal{W}^+} L_r, g_{A_2} = \nabla_{\mathcal{W}^+} L_a, g_{S_2} = \nabla_{\mathcal{W}^+}  L_s,
	\label{eq:style2}
\end{equation}

\noindent where, $L_{r}$, $L_{a}$, and $L_{s}$ are the LPIPS, appearance, and style losses computed between $I_2$ and $I_G$.
To enforce orthogonality, we would like to minimize ${g_{R_2}}^{\transpose} (g_{A_2} + g_{S_2})$.
We achieve this by projecting away the component of $g_{R_2}$ parallel to $(g_{A_2} + g_{S_2})$, using the structure-appearance gradient orthogonalization
\begin{equation}
	g_{R_2} = g_{R_2} - \frac{{g_{R_2}}^{\transpose} (g_{A_2} + g_{S_2})}{{\lVert g_{A_2} + g_{S_2}\rVert}_2^2}(g_{A_2} + g_{S_2})
	\label{eq:style2}
\end{equation}

\noindent after every iteration in stage~$2$ of optimization.

\section{Experiments and Results}

\subsection{Implementation Details}

\textbf{Datasets.} We use the Flickr-Faces-HQ dataset (FFHQ)~\cite{karras2019astylebased} that contains \num{70000} high-quality images of human faces.
Flickr-Faces-HQ has significant variation in terms of ethnicity, age, and hair style patterns.
We select tuples of images $(I_1,I_2,I_3)$ based on the following constraints: (a) each image in the tuple should have at least $18\%$ of pixels contain hair, and (b) $I_1$ and $I_2$'s face regions must align to a certain degree.
To enforce these constraints we extract hair and face masks using the Graphonomy segmentation network~\cite{gong2019graphonomy} and estimate 68 2D facial landmarks using 2D-FAN~\cite{bulat2017far}.
For every $I_1$ and $I_2$, we compute the intersection over union (IoU) and pose distance (PD) using the corresponding face masks, and facial landmarks.
Finally, we distribute selected tuples into three categories, \textit{easy}, \textit{medium}, and \textit{difficult}, such that the following IoU and PD constraints are both met
\begin{table}[!htb]
	\begin{center}
		\begin{tabular}{cccc}
			\toprule
			Category  & Easy         & Medium       & Difficult    \\
			\midrule
			IoU range & $(0.8, 1.0]$ & $(0.7, 0.8]$ & $(0.6, 0.7]$ \\
			PD range  & $[0.0, 2.0)$ & $[2.0, 4.0)$ & $[4.0, 5.0)$ \\
			\bottomrule
		\end{tabular}
		\caption{\textbf{Criteria used to define the alignment of head pose between sample tuples.}}
	\end{center}
\end{table}

\vspace{-.3in}
\textbf{Training Parameters.} We used the Adam optimizer~\cite{kingma2017adam} with an initial learning rate of 0.1 and annealed it using a cosine schedule~\cite{karras2020stylegan2}.
The optimization occurs in two stages, where each stage consists of~\num{1000} iterations.
Based on ablation studies, we selected an appearance loss weight $\lambda_{a}$ of 40, style loss weight $\lambda_{s}$ of~\num{1.5e4}, and noise regularization weight $\lambda_{n}$ of~\num{1e5}. We set the remaining loss weights to 1.

\subsection{Effect of Two-Stage Optimization}

Optimizing all losses in our objective function causes the framework to diverge.
While the identity is reconstructed, the hair transfer fails (Figure~\ref{fig:1-2-stage}).
The structure and shape of the synthesized hair is not preserved, causing undesirable results.
On the other hand, performing optimization in two stages clearly improves the synthesis process leading to generation of photorealistic images that are consistent with the provided references.
Not only is the identity reconstructed, the hair attributes are transferred as per our requirements.

\begin{figure}[!t]
	\begin{center}
		\includegraphics[width=1.0\linewidth]{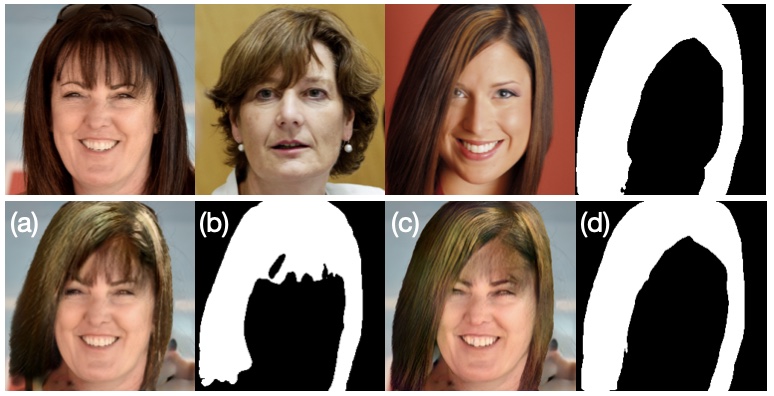}
		\vspace{-.4in}
	\end{center}
	\caption{\textbf{Effect of Gradient Orthogonalization (GO).}
		\textbf{Row 1}: Reference images (from left to right): Identity, target hair appearance and style, target hair structure and shape.
		\textbf{Row 2}: Pairs (a) and (b), and (c) and (d) are synthesized images and their corresponding hair masks for no-GO and GO methods, respectively.}
	\label{fig:long}
	\label{fig:gp_compare}
\end{figure}

\begin{figure}[!t]
	\begin{center}
		\includegraphics[width=1.0\linewidth]{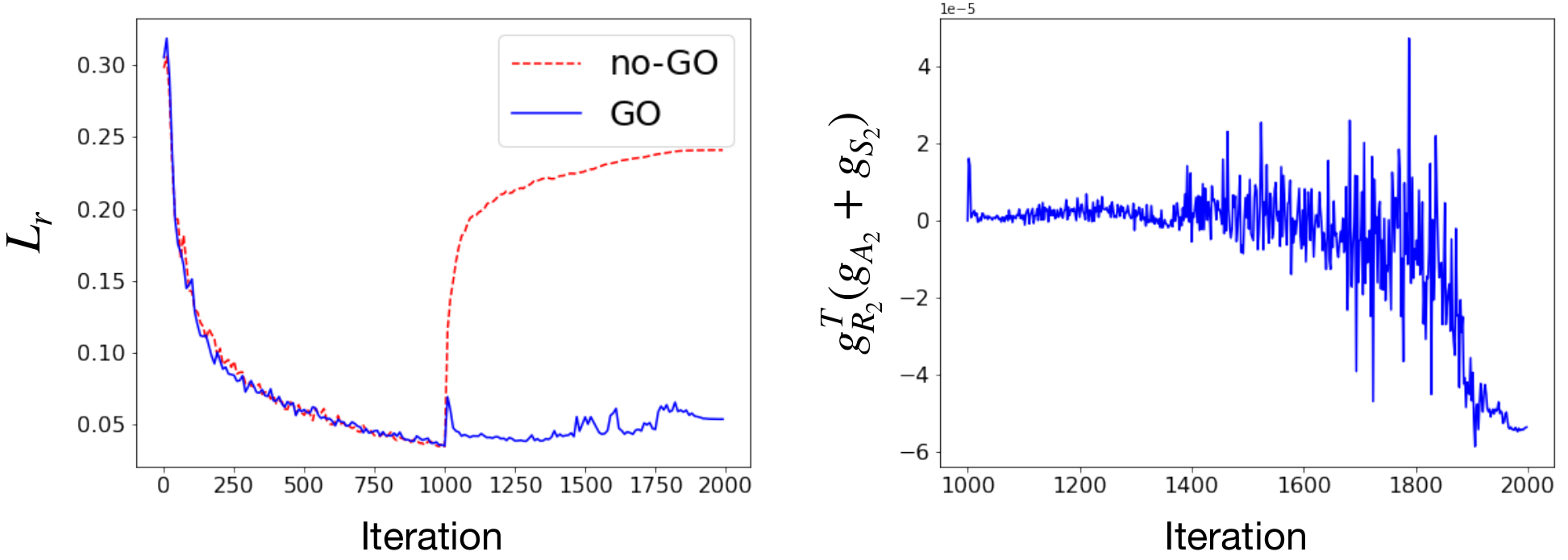}
		\vspace{-.4in}
	\end{center}
	\caption{\textbf{Effect of Gradient Orthogonalization (GO).}
		\textbf{Left}: LPIPS hair reconstruction loss (GO vs no-GO) vs iterations.
		\textbf{Right}: Trend of ${g_{R_2}}^{\transpose} (g_{A_2} + g_{S_2})$ ($\times$1e-5) in stage 2 of optimization.}
	\label{fig:gp_losses}
	\vspace{-0.1in}
\end{figure}

\subsection{Effect of Gradient Orthogonalization}
\label{sec:effect-gradient-orth}

We compare two variations of our framework: no-GO and GO.
GO involves manipulating $L_r$'s gradients via gradient orthogonalization, whereas no-GO keeps $L_r$ untouched.
No-GO is unable to retain the target hair shape, causing $L_r$ to increase in stage 2 of optimization i.e., after iteration 1000 (Figures~\ref{fig:gp_compare} \&~\ref{fig:gp_losses}).
The appearance and style losses, being position invariant, do not contribute to the shape.
GO, on the other hand, uses the reconstruction loss in stage $2$ and retains the target hair shape.
As a result, the IoU computed between $M^h_2$ and $M^h_G$ increases from $0.857$ (for no-GO) to $0.932$ (GO).

In terms of gradient disentanglement, the similarity between $g_{R_2}$ and $(g_{A_2} + g_{S_2})$ decreases with time, indicating that our framework is able to disentangle person 2's hair shape from its appearance and style (Figure~\ref{fig:gp_losses}).
This disentanglement allows a seamless transfer of person 3's hair appearance and style to the synthesized image without causing model divergence.
Here on, we will use the GO version of our framework for comparisons and analysis.

\subsection{Comparison with SOTA}

\textbf{Hair Style Transfer.} We compare our approach with the SOTA model MichiGAN.
MichiGAN contains separate modules to estimate: (1) hair appearance, (2) hair shape and structure, and (3) background.
The appearance module bootstraps the generator with its output feature map, replacing the randomly sampled latent code in traditional GANs~\cite{goodfellow2014gan}.
The shape and structure module outputs hair masks and orientation masks, denormalizing each SPADE ResBlk~\cite{park2019spade} in the backbone generation network.
Finally, the background module progressively blends the generator outputs with background information.
In terms of training, MichiGAN follows the pseudo-supervised regime.
Specifically, the features, that are estimated by the modules, from the same image are fed into MichiGAN to reconstruct the original image.
At test time, FID is calculated for 5000 images at 512 px resolution uniform randomly sampled from FFHQ's test split.

\begin{figure}[!t]
	\begin{center}
		\includegraphics[width=1.0\linewidth]{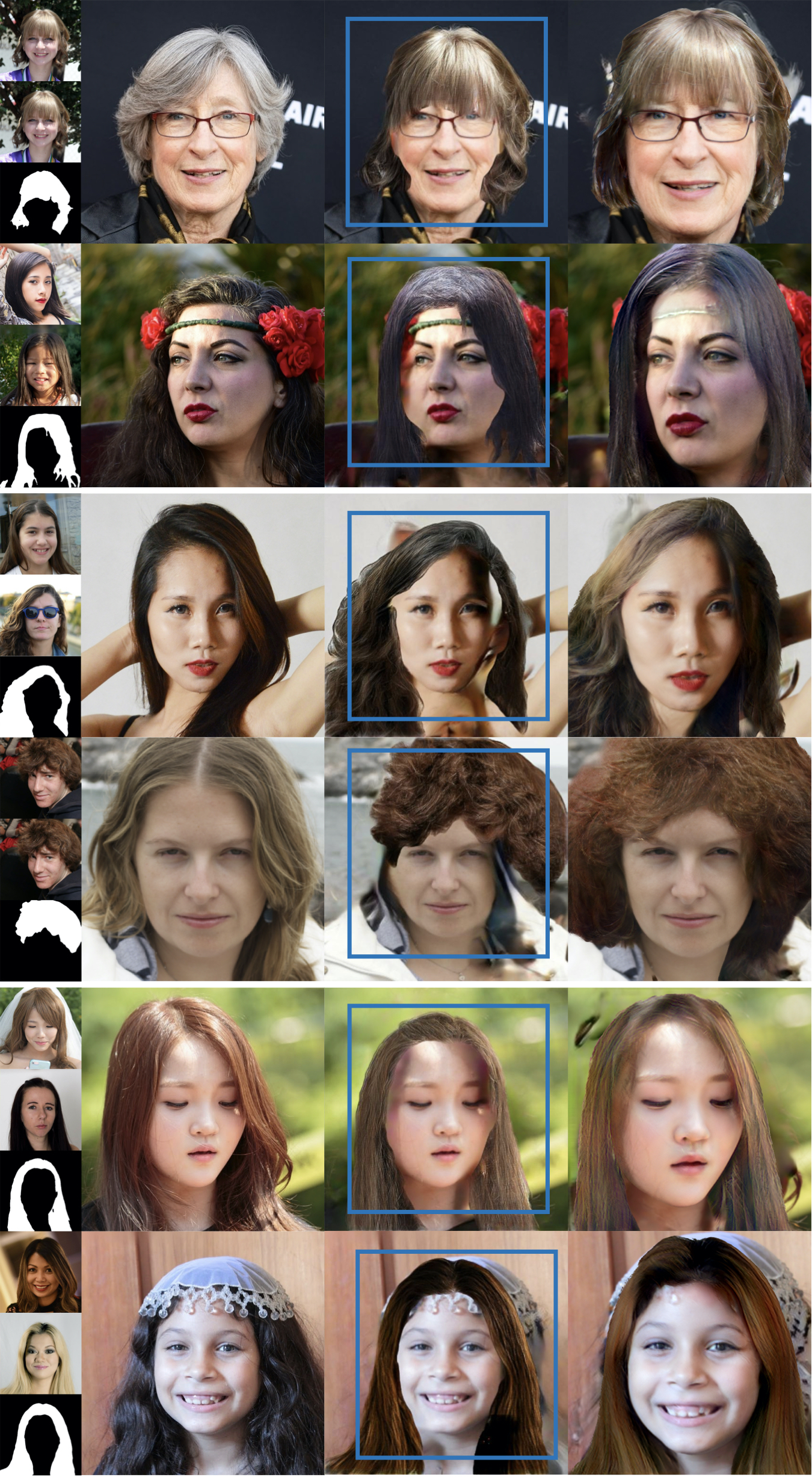}
		\vspace{-.4in}
	\end{center}
	\caption{Qualitative comparison of MichiGAN and LOHO.
		\textbf{Col 1 (narrow)}: Reference images.
		\textbf{Col 2}: Identity person
		\textbf{Col 3}: MichiGAN output.
		\textbf{Col 4}: LOHO output (zoomed in for better visual comparison).
		\textbf{Rows 1-2}: MichiGAN ``copy-pastes'' the target hair attributes while LOHO blends the attributes, thereby synthesizing more realistic images.
		\textbf{Rows 3-4}: LOHO handles misaligned examples better than MichiGAN.
		\textbf{Rows 5-6}: LOHO transfers the right style information.}
	\label{fig:michigan_compare}
	\vspace{-0.1in}
\end{figure}

To ensure that our results are comparable, we follow the above procedure and compute FID scores~\cite{heusel2017fid} for LOHO.
In addition to computing FID on the entire image, we calculate the score solely relying on the synthesized hair and facial regions with the background masked out.
Achieving a low FID score on masked images would mean that our model is indeed capable of synthesizing realistic hair and face regions.
We call this LOHO-HF.
As MichiGAN's background inpainter module is not publicly available, we use GatedConv~\cite{yu2019gatedconv} to inpaint relevant features in the masked out hair regions.

Quantitatively, LOHO outperforms MichiGAN.
Our method achieves an FID score of 8.419, while MichiGAN achieves 10.697 (Table~\ref{tab:fid}).
This improvement indicates that our optimization framework is able to synthesize high quality images.
LOHO -HF achieves an even lower score of $4.847$, attesting to the superior quality of the synthesized hair and face regions.

Qualitatively, our method is able to synthesize better results for challenging examples.
LOHO naturally blends the target hair attributes with the target face (Figure~\ref{fig:michigan_compare}).
MichiGAN na\"ively copies the target hair on the target face, causing lighting inconsistencies between the two regions.
LOHO handles pairs with varying degrees of misalignment whereas MichiGAN is unable to do so due
to its reliance on blending background and foreground information in pixel
space rather than latent space.
Lastly, LOHO transfers relevant style information, on par with MichiGAN.
In fact, due to our addition of the style objective to optimize second-order
statistics by matching Gram matrices, LOHO synthesizes hair with varying colour
even when the hair shape source person has uniform hair colour, as in the
bottom two rows of Figure~\ref{fig:michigan_compare}.

\begin{table}[!t]
	\begin{center}
		\begin{tabular}{cccc}
			\toprule
			Method             & MichiGAN & LOHO-HF & LOHO  \\
			\midrule
			FID $(\downarrow)$ & 10.697   & 4.847   & 8.419 \\
			\bottomrule
		\end{tabular}
		\vspace{-.2in}
	\end{center}
	\caption{\textbf{Frechet Inception Distance (FID) for different methods}. We use~\num{5000} images uniform-randomly sampled from the testing set of FFHQ. $\downarrow$ indicates that lower is better.}
	\label{tab:fid}
\end{table}

\begin{table}[!t]
	\begin{center}
		\begin{tabular}{cccc}
			\toprule
			Method                                    & I2S            & I2S++   & LOHO            \\
			\midrule
			PSNR (dB) $(\uparrow)$                    & -              & $22.48$ & $32.2\pm 2.8$   \\
			SSIM $(\uparrow)$                         & -              & $0.91$  & $0.93 \pm 0.02$ \\
			\midrule
			${\lVert w^{*} - \hat{\mathbf{w}}\rVert}$ & $[30.6, 40.5]$ & -       & $37.9 \pm 3.0$  \\
			\bottomrule
		\end{tabular}
		\vspace{-.2in}
	\end{center}
	\caption{\textbf{PSNR, SSIM and range of acceptable latent distances ${\lVert w^{*} - \hat{\mathbf{w}}\rVert}$ for different methods}. We use randomly sampled 5000 images from the testing set of FFHQ. - indicates N/A. $\uparrow$ indicates that higher is better.}
	\label{tab:psnr}
	\vspace{-0.1in}
\end{table}

\textbf{Identity Reconstruction Quality.} We also compare LOHO with two recent image embedding methods: I2S~\cite{abdal2019image2stylegan} and I2S++~\cite{abdal2020i2s++}.
introduces the framework that is able to reconstruct images of high quality by optimizing the $\mathcal{W}^+$ latent space.
I2S also shows how the latent distance, calculated between the optimized style latent code $w^*$ and $\hat{\mathbf{w}}$ of the average face, is related to the quality of synthesized images.
I2S++, additionally to I2S, optimizes the noise space $\mathcal{N}$ in order to reconstruct images with high PSNR and SSIM values.
Therefore, to assess LOHO's ability to reconstruct the target identity with high quality, we compute similar metrics on the facial region of synthesized images.
Since inpainting in latent space is an integral part of LOHO we compare our results with I2S++'s performance on image inpainting at 512 px resolution.

Our model, despite performing the difficult task of hair style transfer, is able to achieve comparable results (Table~\ref{tab:psnr}).
I2S shows that the acceptable latent distance for a valid human face is in $[30.6, 40.5]$ and LOHO lies within that range.
Additionally, our PSNR and SSIM scores are better than I2S++, proving that LOHO reconstructs identities that satisfy local structure information.

\begin{figure}[t]
	\begin{center}
		\includegraphics[width=1.0\linewidth]{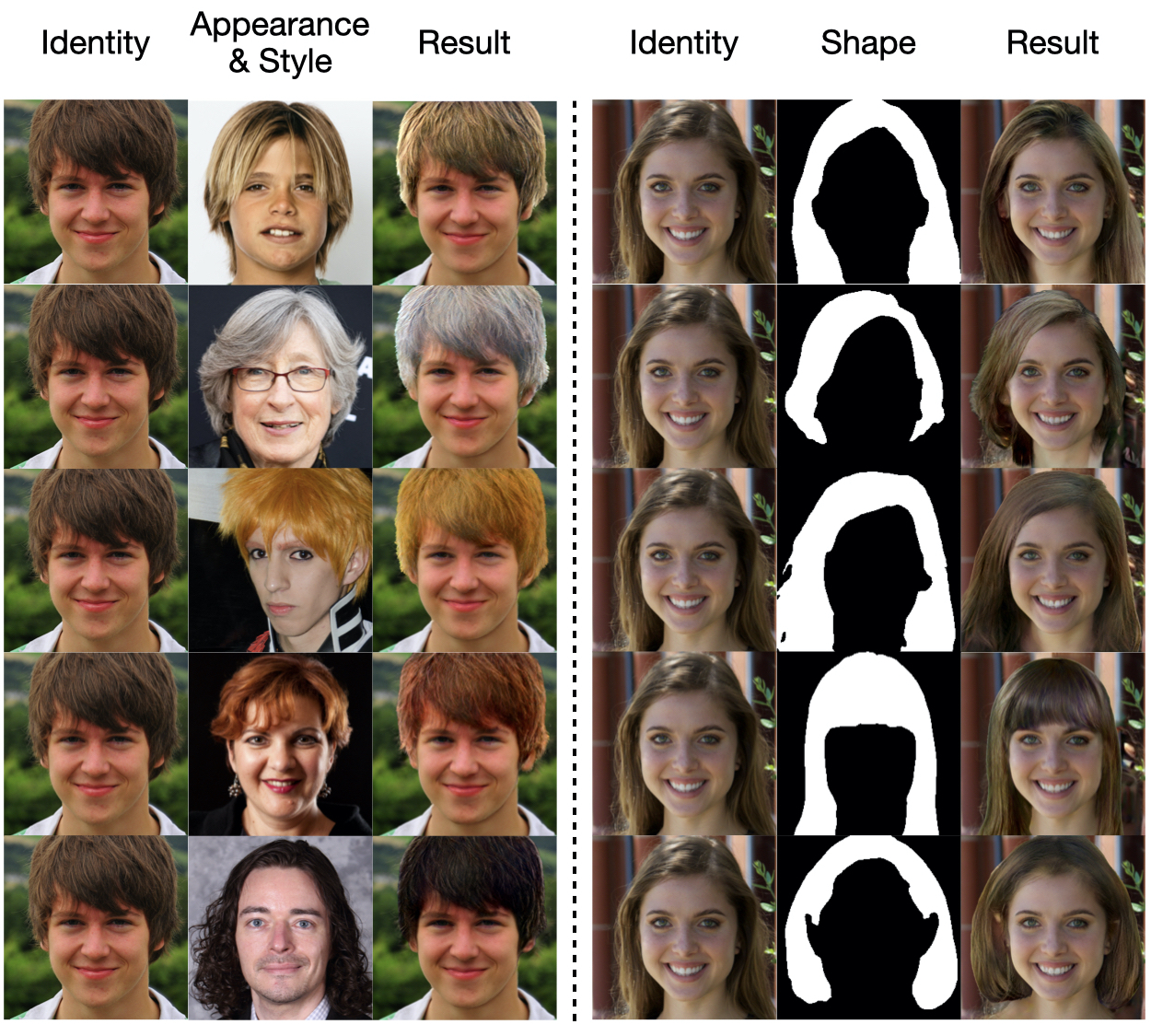}
		\vspace{-.4in}
	\end{center}
	\caption{\textbf{Individual attribute editing}. The results show that our model is able to edit individual hair attributes (\textbf{left}: appearance \& style left, \textbf{right}: shape) without them interfering with each other.}
	\label{fig:appearance_shape}
	\vspace{-0.1in}
\end{figure}

\subsection{Editing Attributes}

\begin{figure}[t]
	\begin{center}
		\includegraphics[width=1.0\linewidth]{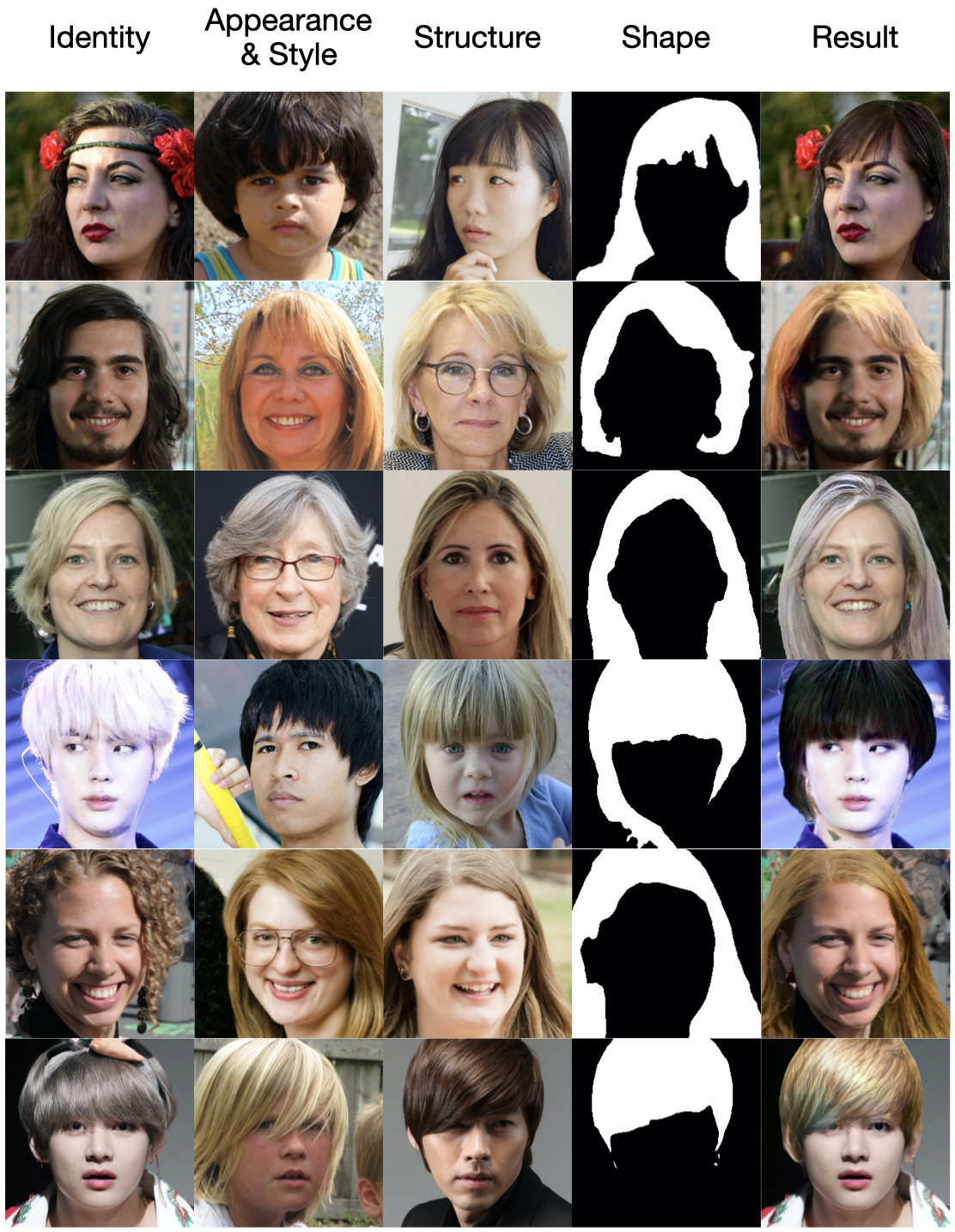}
		\vspace{-.4in}
	\end{center}
	\caption{\textbf{Multiple attributes editing}. The results show that our model is able to edit hair attributes jointly without the interference of each other.}
	\label{fig:multiple_attributes}
	\vspace{-0.1in}
\end{figure}

Our method is capable of editing attributes of in-the-wild portrait images.
In this setting, an image is selected and then an attribute is edited individually by providing reference images.
For example, the hair structure and shape can be changed while keeping the hair appearance and background unedited.
Our framework computes the non-overlapping hair regions and infills the space with relevant background details.
Following the optimization process, the synthesized image is blended with the inpainted background image.
The same holds for changing the hair appearance and style.
LOHO disentangles hair attributes and allows editing them individually and jointly, thereby leading to desirable results (Figures~\ref{fig:appearance_shape} \&~\ref{fig:multiple_attributes}) .

\section{Limitations}

\begin{figure}[t]
	\begin{center}
		\includegraphics[width=1.0\linewidth]{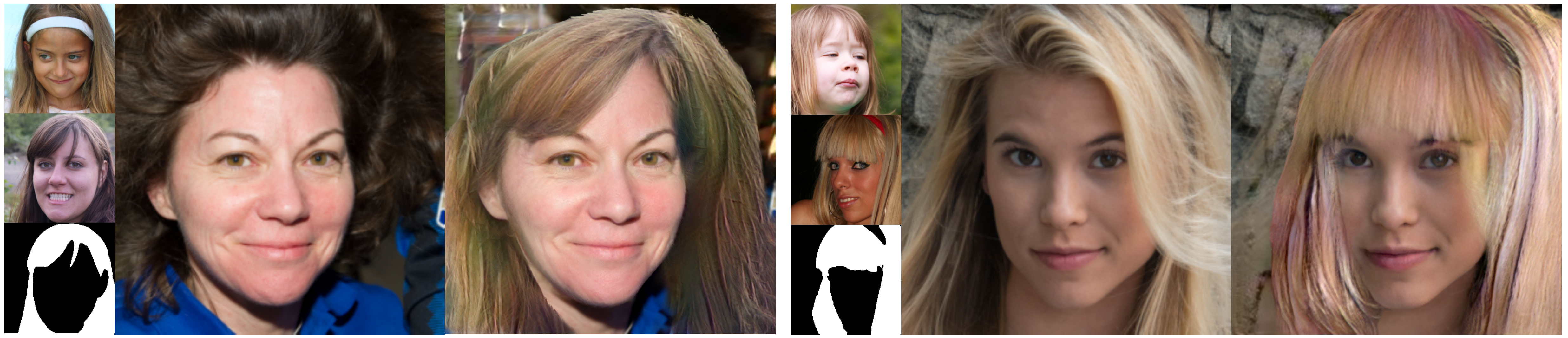}
		\vspace{-.4in}
	\end{center}
	\caption{\textbf{Misalignment examples.}
		\textbf{Col 1 (narrow)}: Reference images.
		\textbf{Col 2}: Identity image.
		\textbf{Col 3}: Synthesized image.
		Extreme cases of misalignment can result in misplaced hair.}
	\label{fig:misalignment}
	\vspace{-0.1in}
\end{figure}

Our approach is susceptible to extreme cases of misalignment (Figure~\ref{fig:misalignment}).
In our study, we categorize such cases as \textit{difficult}.
They can cause our framework to synthesize unnatural hair shape and structure.
GAN based alignment networks~\cite{zakharov2019fewshot, burkov2020neuralhead} may be used to transfer pose, or alignment of hair across \textit{difficult} samples.

\begin{figure}[t]
	\begin{center}
		\includegraphics[width=1.0\linewidth]{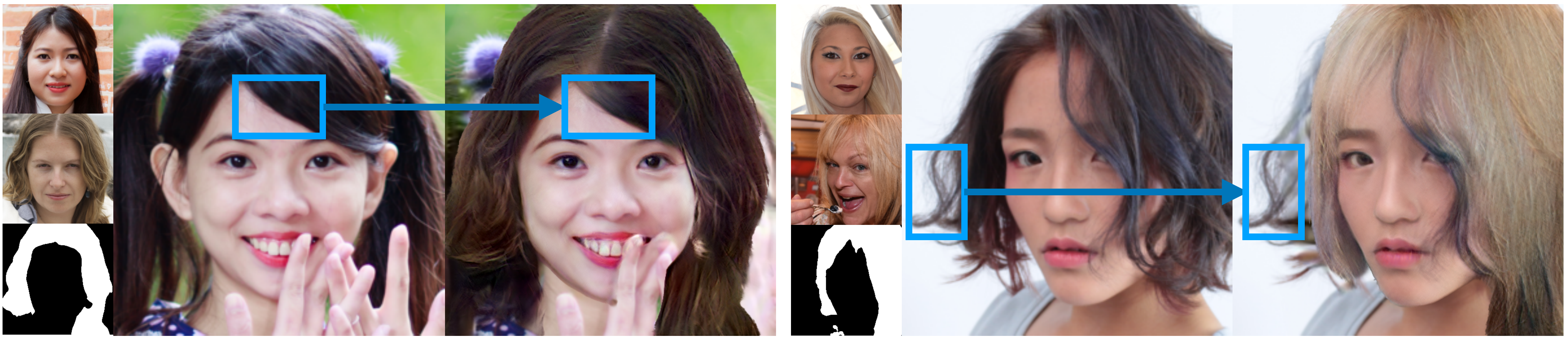}
		\vspace{-.4in}
	\end{center}
	\caption{\textbf{Hair trail}.
		\textbf{Col 1 (narrow)}: Reference images.
		\textbf{Col 2}: Identity image.
		\textbf{Col 3}: Synthesized image.
		Cases where there are remnants of hair information from the identity person.
		The regions marked inside the blue box carries over to the synthesized image.}
	\label{fig:trail}
\end{figure}

In some examples, our approach can carry over hair details from the identity person (Figure~\ref{fig:trail}).
This can be due to Graphonomy~\cite{gong2019graphonomy}'s imperfect segmentation of hair.
More sophisticated segmentation networks~\cite{yuan2020ocrseg, tao2020hierarchicalMA} can be used to mitigate this issue.

\section{Conclusion}

Our introduction of LOHO, an optimization framework that performs hairstyle
transfer on portrait images, takes a step in the direction of
spatially-dependent attribute manipulation with pretrained GANs.
We show that developing algorithms that approach specific synthesis tasks,
such as hairstyle transfer, by manipulating the latent space of expressive
models trained on more general tasks, such as face synthesis, is effective for
completing many downstream tasks without collecting large training datasets.
GAN inversion approach is able to solve problems such as
realistic hole-filling more effectively, even than feedforward GAN pipelines
that have access to large training datasets.
There are many possible improvements to our approach for hair synthesis, such
as introducing a deformation objective to enforce alignment over a wide range
of head poses and hair shapes, and improving convergence by predicting an
initialization point for the optimization process.

\clearpage

{\small
	\bibliographystyle{ieee_fullname}
	\bibliography{hair_style_transfer}
}

\clearpage

\appendix
\section{Images and Masks}

\begin{figure}[t]
	\begin{center}
		\includegraphics[width=1.0\linewidth]{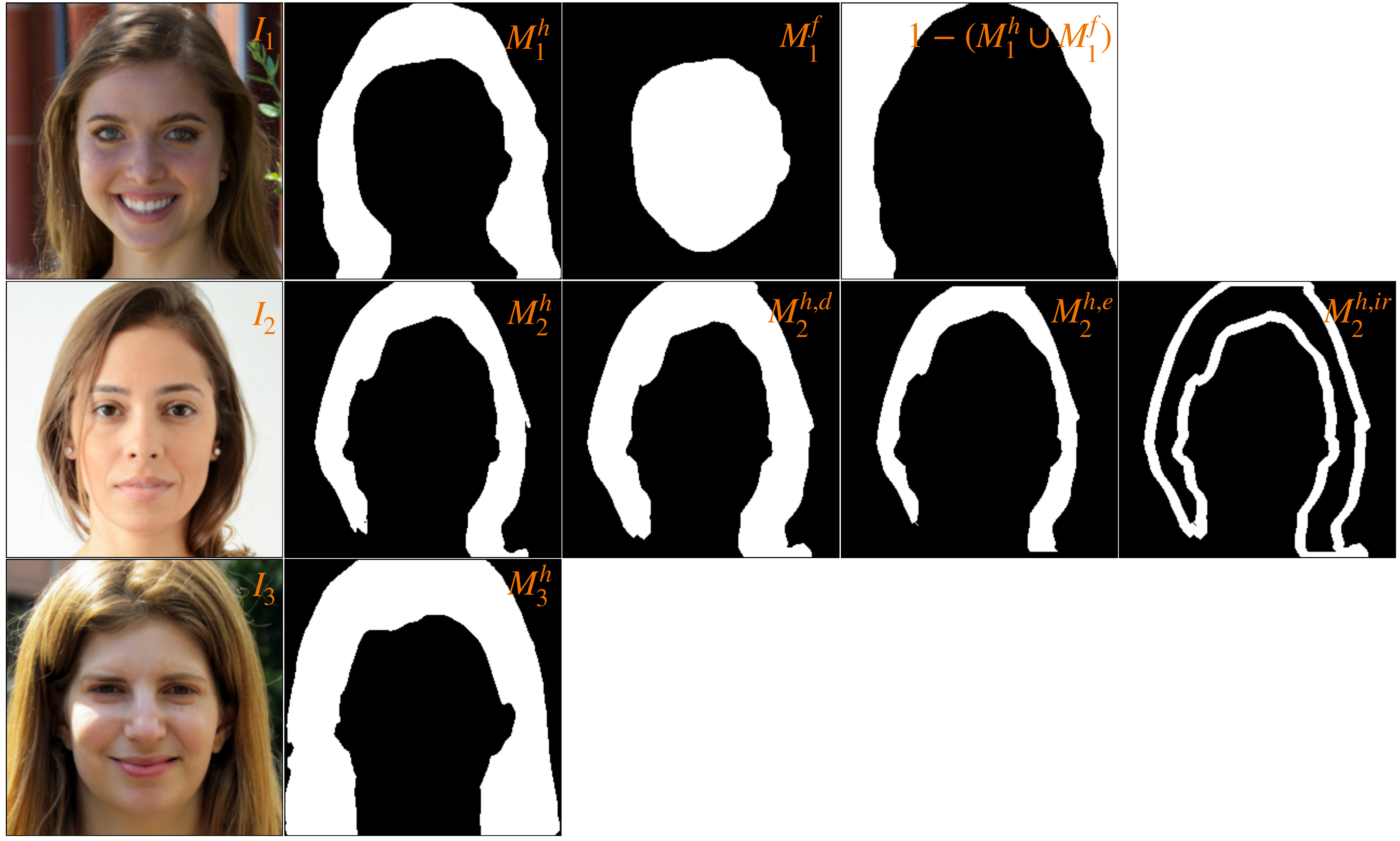}
		\vspace{-.3in}
	\end{center}
	\caption{Tuple $(I_1, I_2, I_3)$ and relevant masks used in LOHO.}
	\label{fig:long}
	\label{fig:images_and_masks}
\end{figure}

\noindent For each selected tuple $(I_1, I_2, I_3)$, we extract hair and face masks using Graphonomy~\cite{gong2019graphonomy}.
We separately dilate and erode $M^h_2$, the hair mask of $I_2$, to produce the dilated version, $M^{h,d}_2$, and the eroded version, $M^{h,e}_2$.
Using $M^{h,d}_2$ and $M^{h,e}_2$, we compute the ignore region $M^{h,ir}_2$.
We exclude the ignore region from the background and let StyleGANv2 inpaint relevant features.
We want to optimize for reconstruction of $I_1$'s face, reconstruction of $I_2$'s hair shape and structure, transfer of $I_3$'s hair appearance and style, and inpainting of the ignore region.
Given a tuple, Figure~\ref{fig:images_and_masks} shows the images and relevant masks used during optimization.

\section{Alignment Metrics}

\noindent To categorize each selected tuple $(I_1, I_2, I_3)$, we calculate the Intersection over Union (IoU) and pose distance (PD) between face masks, and 68 2D facial landmarks.
We extract the masks using Graphonomy~\cite{gong2019graphonomy}, and estimate landmarks using 2D-FAN~\cite{bulat2017far}.

IoU and PD quantify to what degree two faces align.
Given the two binary face masks, $M^f_1$ and $M^f_2$, we compute IoU as
\begin{equation}
	\text{IoU} = \frac{M^f_1 \cap M^f_2}{M^f_1 \cup M^f_2}.
	\label{eqn:iou}
\end{equation}

\begin{figure}[t]
	\begin{center}
		\includegraphics[width=1.0\linewidth]{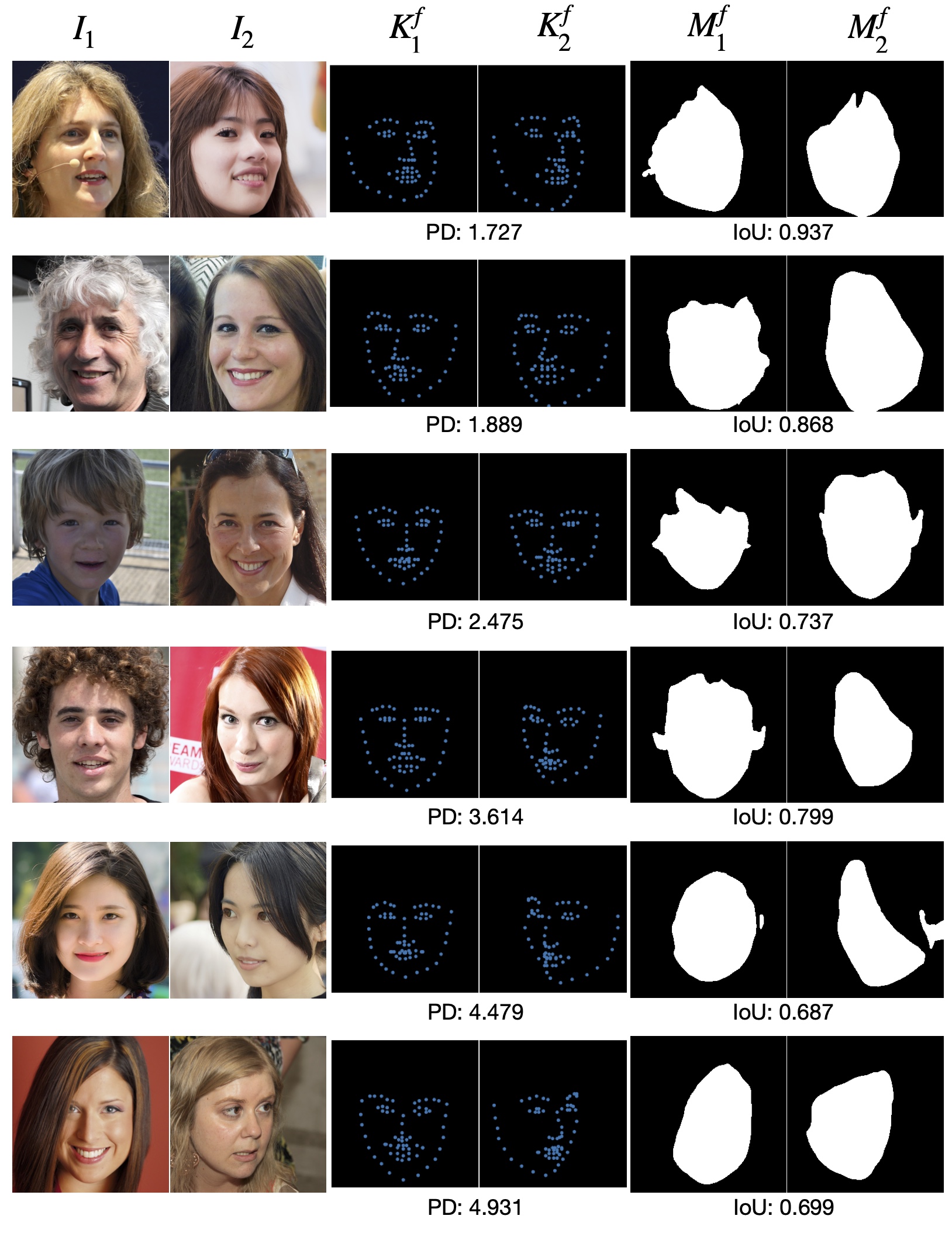}
		\vspace{-.3in}
	\end{center}
	\caption{IoU and PD for tuples in each category.
		\textbf{Rows 1-2}: \textit{Easy} tuples.
		\textbf{Rows 3-4}: \textit{Medium} tuples.
		\textbf{Rows 5-6}: \textit{Difficult} tuples.}
	\label{fig:long}
	\label{fig:alignment_samples}
\end{figure}

\noindent The pose distance (PD), on the other hand, is defined in terms of facial landmarks.
Given the two 68 2D facial landmarks, $K^f_1 \in \mathbb{R}^{68 \times 2}$ and $K^f_2 \in \mathbb{R}^{68 \times 2}$, corresponding to $I_1$ and $I_2$, PD is calculated by averaging the $L_2$ distances computed between each landmark
\begin{equation}
	\text{PD} = \frac{1}{68} \sum_{k=1}^{68} {\lVert K^f_{1,k} - K^f_{2,k} \rVert}_2
	\label{eqn:pd}
\end{equation}

\noindent where $k$ indexes the 2D landmarks.
Therefore, a tuple where $I_1$ and $I_2$ are the same person (Figure~\ref{fig:alignment_samples}) would have an IoU of 1.0 and PD of 0.0.

\section{StyleGANv2 Architecture}

\noindent StyleGANv2~\cite{karras2020styleganv2} can synthesize novel photorealistic images at different resolutions including $128^2$, $256^2$, $512^2$ and $1024^2$.
The number of layers in the architecture therefore depends on the resolution of images being synthesized.
Additionally, the size of the extended latent space $\mathcal{W}^+$ and the noise space $\mathcal{N}$ also depend on on the resolution.
Embeddings sampled from $\mathcal{W}^+$ are concatenations of 512-dimensional vectors $w$, where $w \in \mathcal{W}^+$.
As our experiments synthesize images of resolution ${512}^2$, the latent space is a vector subspace of~$\mathbb{R}^{15\times 512}$, i.e., $\mathcal{W}^+ \subset \mathbb{R}^{15 \times 512}$.
Additionally, noise maps sampled from $\mathcal{N}$ are tensors of dimension ~$\mathbb{R}^{1\times 1 \times h \times w}$, where $h$ and $w$ match the spatial resolution of feature maps at every layer of the StyleGANv2 generator.

\section{Effect of Regularizing Noise Maps}

\begin{figure}[t]
	\begin{center}
		\includegraphics[width=1.0\linewidth]{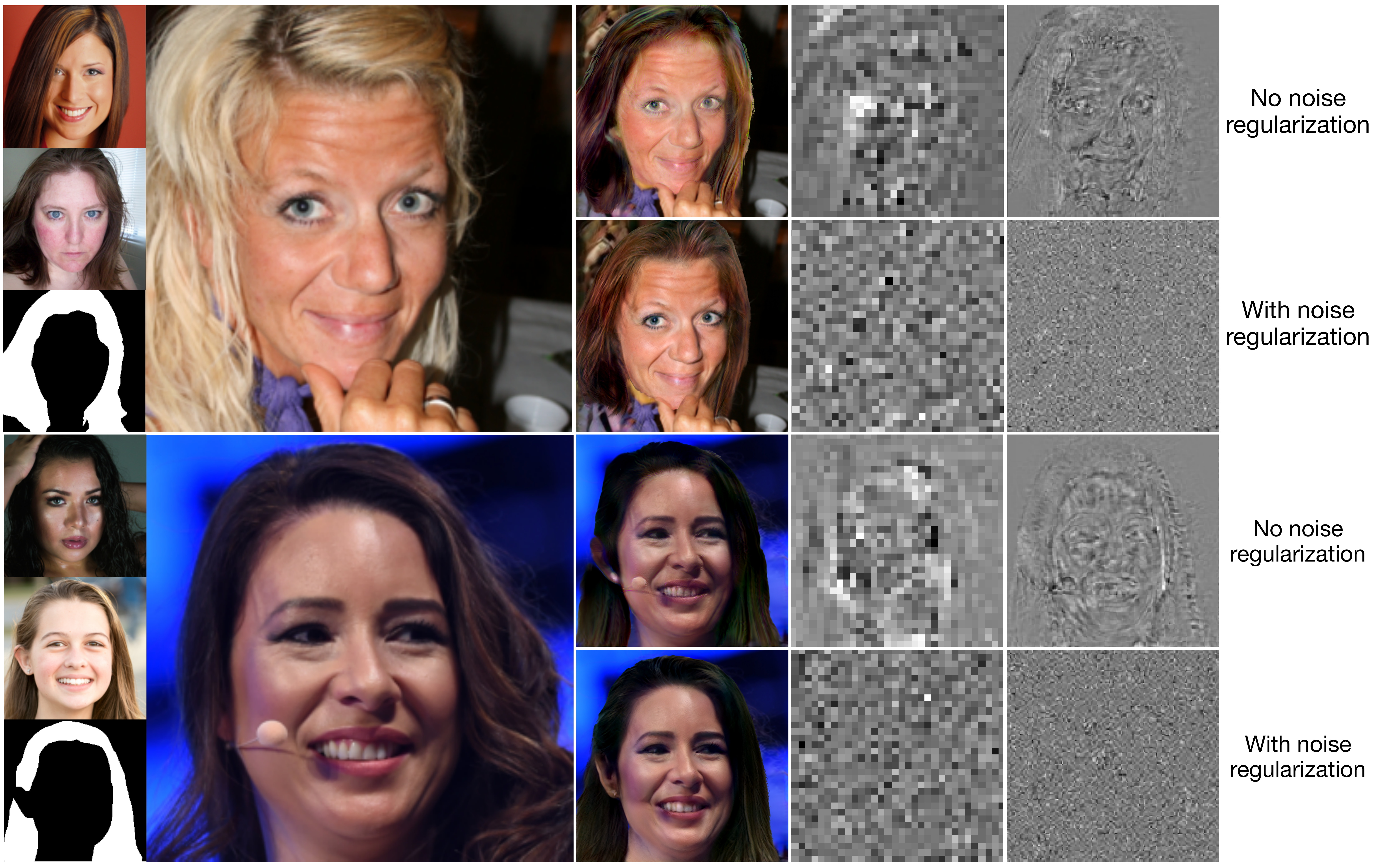}
		\vspace{-.3in}
	\end{center}
	\caption{\textbf{Effect of regularizing noise maps}.
		\textbf{Col 1 (narrow)}: Reference images.
		\textbf{Col 2}: Identity person.
		\textbf{Col 3}: Synthesized images.
		\textbf{Cols 4\&5}: Noise maps at different resolutions.}
	\label{fig:long}
	\label{fig:noise_compare}
\end{figure}

\noindent To understand the effect of noise map regularization, we visualize noise maps at different resolutions post optimization.
When the regularization term is set to zero, we normalize the noise maps to be zero mean and unit variance.
This causes the optimization to inject actual signal into the noise maps, thereby causing overfitting.
Figure~\ref{fig:noise_compare} shows that the noise maps encode structural information of the facial region, which is not desirable, and cause the synthesized images to have artifacts in the face and hair regions.
Enabling noise regularization prevents this.

\begin{figure}[t]
	\begin{center}
		\includegraphics[width=1.0\linewidth]{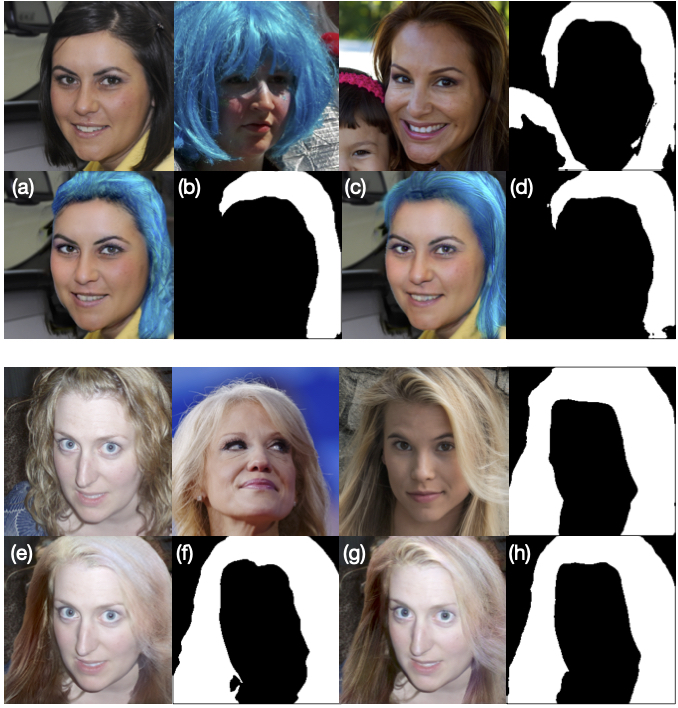}
		\vspace{-.3in}
	\end{center}
	\caption{\textbf{Effect of Gradient Orthogonalization (GO).}
		\textbf{Rows 1\&3}: Reference images (from left to right): Identity, target hair appearance and style, target hair structure and shape.
		\textbf{Rows 2\&4}: Pairs (a) and (b), and (c) and (d) are synthesized images and their corresponding hair masks for no-GO and GO methods, respectively.
		The same holds for pairs (e) and (f), and (g) and (h).}
	\label{fig:long}
	\label{fig:gpnogp}
\end{figure}

\section{Additional Examples of Gradient Orthogonalization}

\noindent Gradient Orthogonalization (GO) allows LOHO to retain the target hair shape and structure during stage 2 of optimization.
Figure~\ref{fig:gpnogp} shows that no-GO fails to maintain the perceptual structure.
On the other hand, GO is able to maintain the target perceptual structure while transferring the target hair appearance and style.
As a result, the IoU calculated between $M^h_2$ and $M^h_G$ increases from 0.547 (no-GO, Figure~\ref{fig:gpnogp} (b)) to 0.603 (GO, Figure~\ref{fig:gpnogp} (d)).
In the same way, the IoU increases from 0.834 (no-GO, Figure~\ref{fig:gpnogp} (f)) to 0.857 (GO, Figure~\ref{fig:gpnogp} (h)).

\begin{figure}[t]
	\begin{center}
		\includegraphics[width=1.0\linewidth]{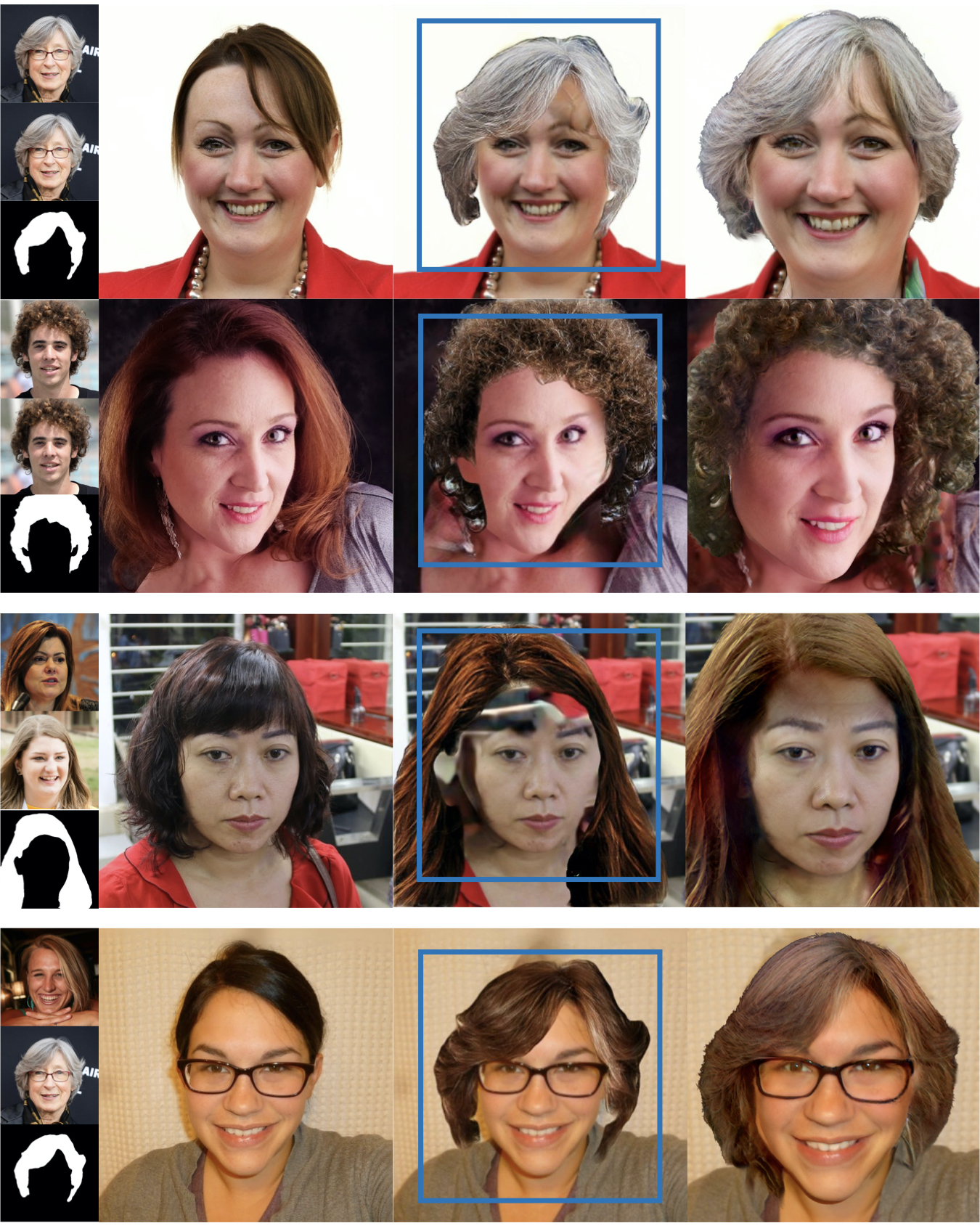}
		\vspace{-.3in}
	\end{center}
	\caption{Qualitative comparison of MichiGAN and LOHO.
		\textbf{Col 1 (narrow)}: Reference images.
		\textbf{Col 2}: Identity person
		\textbf{Col 3}: MichiGAN output.
		\textbf{Col 4}: LOHO output (zoomed in for better visual comparison).
		\textbf{Rows 1-2}: MichiGAN ``copy-pastes'' the target hair attributes while LOHO blends the attributes, thereby synthesizing more realistic images.
		\textbf{Row 3}: LOHO handles misaligned examples better than MichiGAN.
		\textbf{Row 4}: LOHO transfers the right style information.}
	\label{fig:long}
	\label{fig:michigan_compare}
\end{figure}

\section{Additional comparisons with MichiGAN}

\noindent We provide additional evidence to show that LOHO addresses blending and misalignment better than MichiGAN~\cite{tan2020michigan}.
The ignore region $M^{h,ir}_2$ (Figure~\ref{fig:images_and_masks}), in addition to StyleGANv2's powerful learned representations, is able to inpaint relevant hair and face pixels.
This infilling causes the synthesized image to look more photorealistic as compared with MichiGAN.
In terms of style transfer, LOHO achieves similar performance as MichiGAN (Figure~\ref{fig:michigan_compare}).

\section{Additional Results of LOHO}

\begin{figure*}
	\centering
	\includegraphics[width=\linewidth]{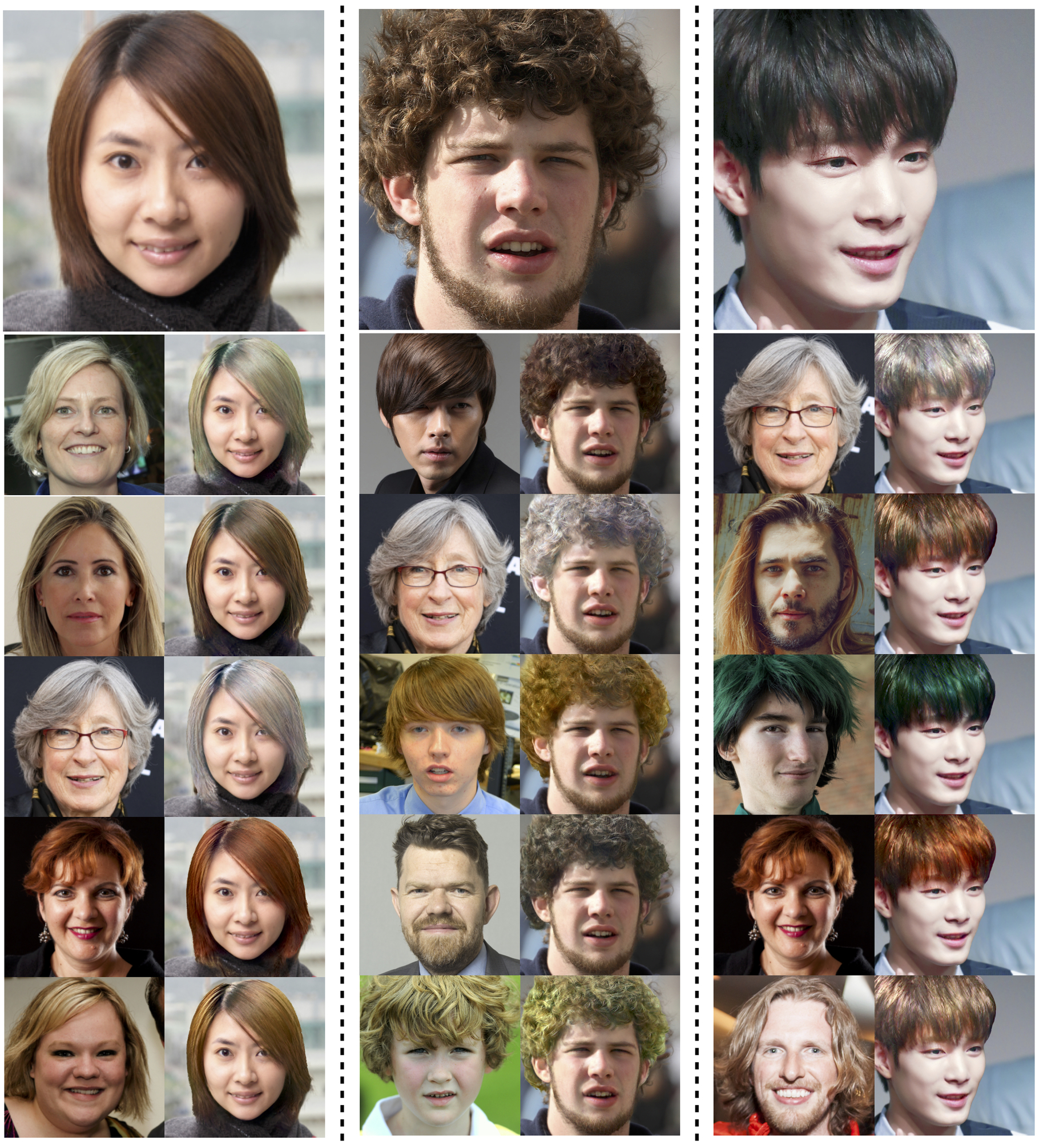}
	\caption{\textbf{Transfer of appearance and style}.
		Given an identity image, and reference image, LOHO transfers the target hair appearance and style while preserving the hair structure and shape.
		\textbf{Row 1}: Identity images.
		\textbf{Rows 2-6}: Hair appearance and style references (Cols: $1,3,5$), and synthesized images (Cols: $2,4,6$).}
	\label{fig:supp_appearance_and_style}
\end{figure*}

\begin{figure*}
	\centering
	\includegraphics[width=\linewidth]{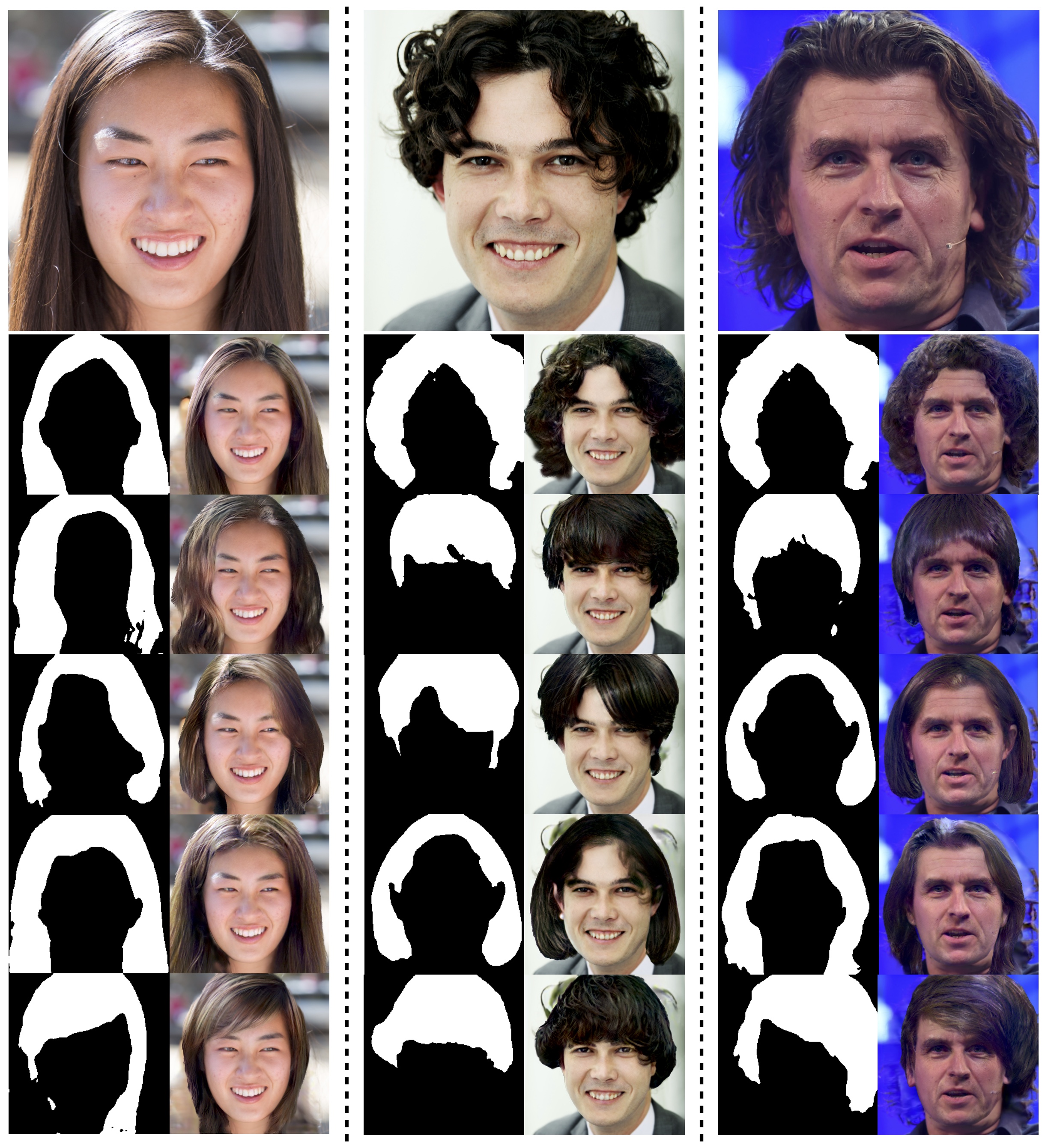}
	\caption{\textbf{Transfer of shape}.
		Given an identity image, and reference image, LOHO transfers the target hair shape while preserving the hair appearance and style.
		\textbf{Row 1}: Identity images.
		\textbf{Rows 2-6}: Hair shape references (Cols: $1,3,5$), and synthesized images (Cols: $2,4,6$).}
	\label{fig:supp_shape}
\end{figure*}

\begin{figure*}
	\centering
	\includegraphics[width=\linewidth]{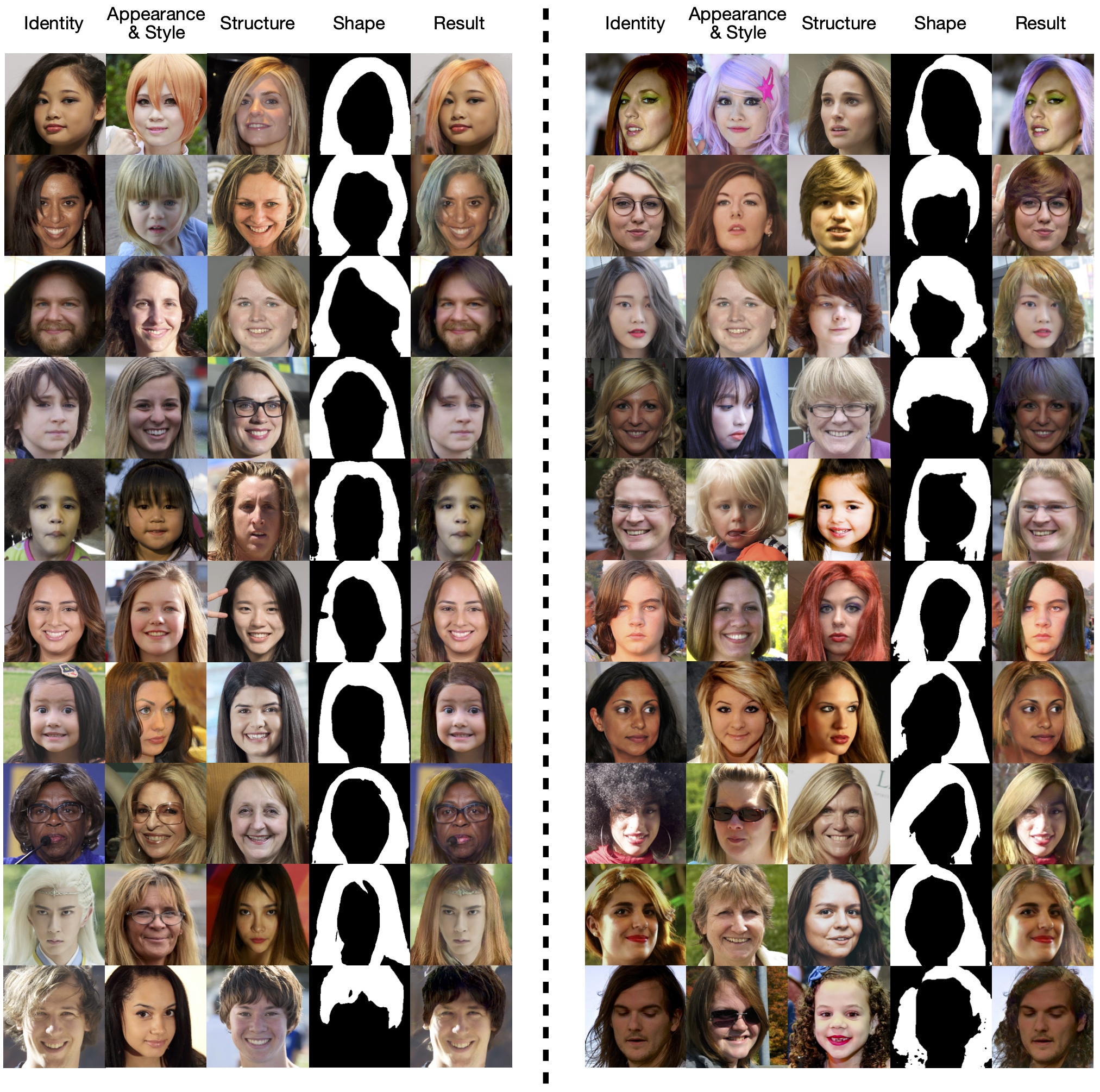}
	\caption{\textbf{Multiple attributes editing}.
		Given an identity image, and reference images, LOHO transfers the target hair attributes.}
	\label{fig:supp_mult_attr_1}
\end{figure*}

\begin{figure*}
	\centering
	\includegraphics[width=\linewidth]{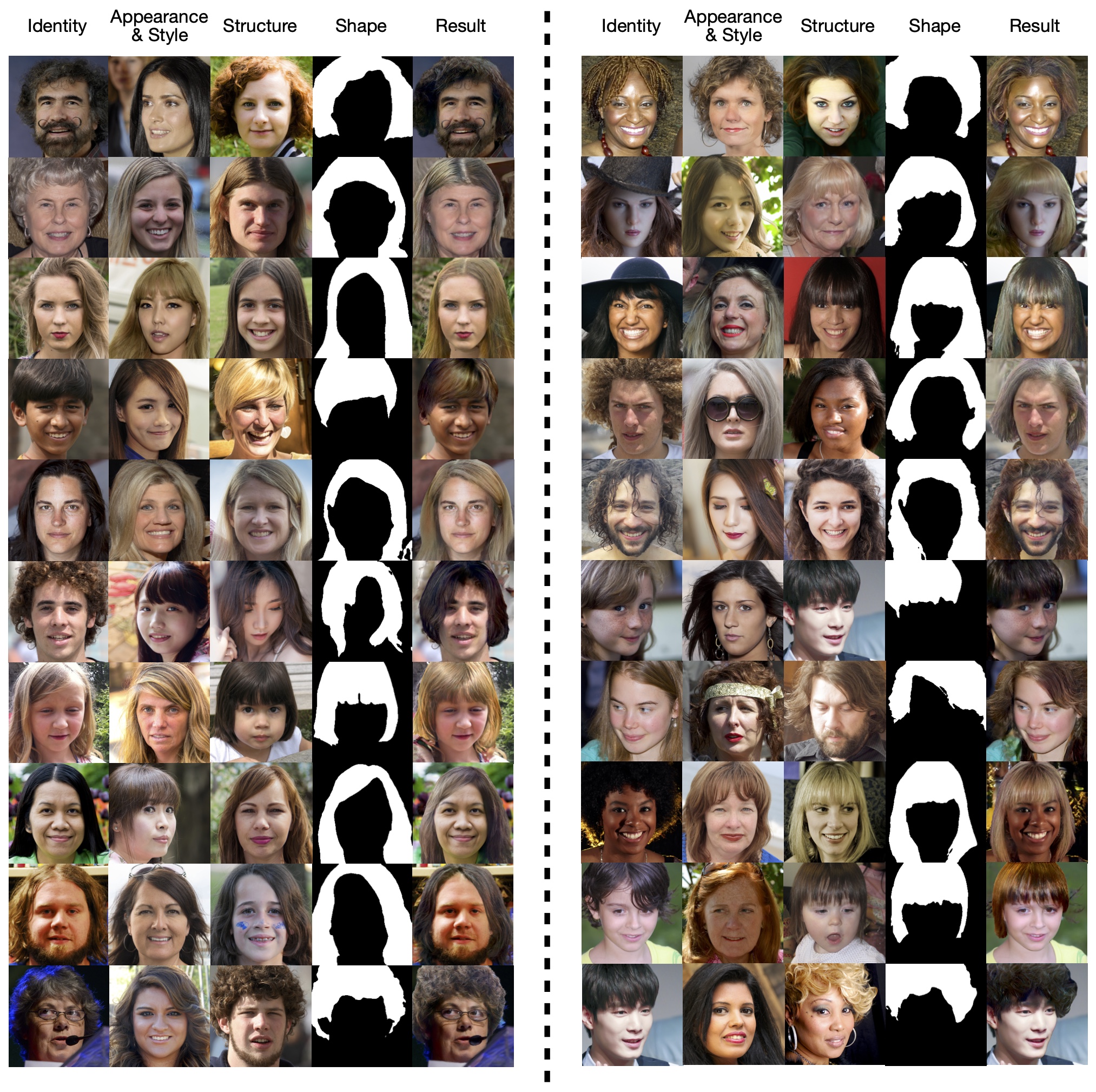}
	\caption{\textbf{Multiple attributes editing}.
		Given an identity image, and reference images, LOHO transfers the target hair attributes.}
	\label{fig:supp_mult_attr_2}
\end{figure*}

\begin{figure*}
	\centering
	\includegraphics[width=\linewidth]{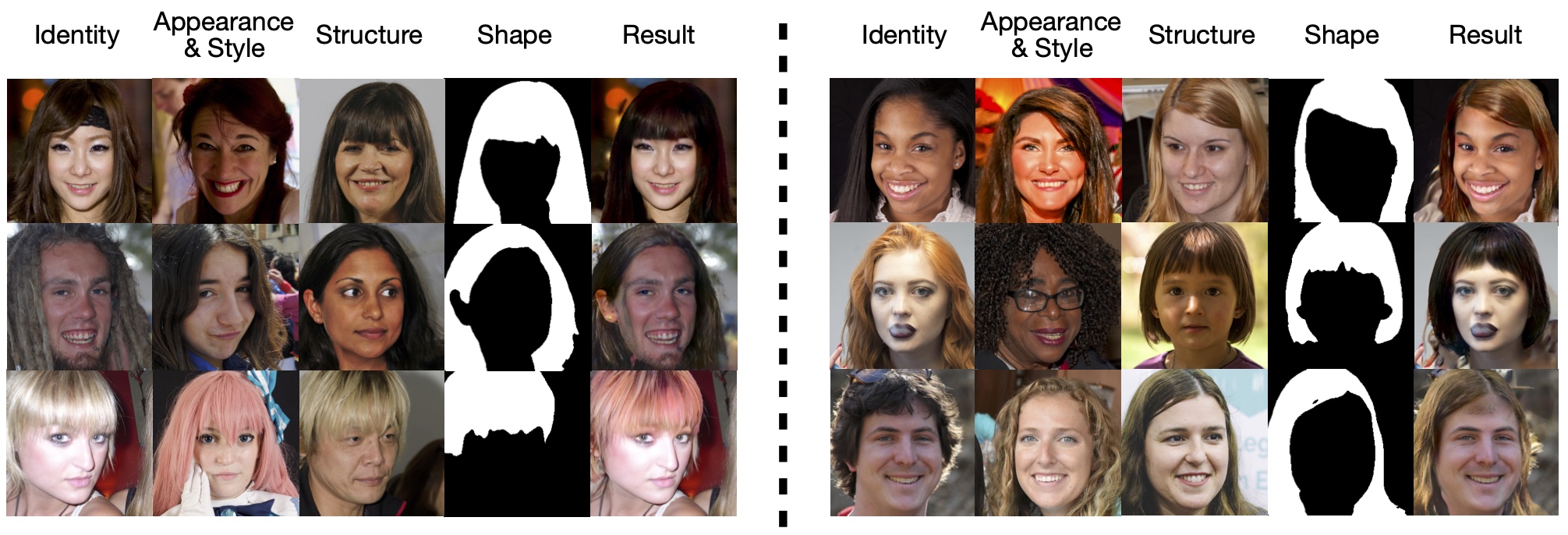}
	\caption{\textbf{Multiple attributes editing}.
		Given an identity image, and reference images, LOHO transfers the target hair attributes.}
	\label{fig:supp_mult_attr_3}
\end{figure*}

\noindent We present results to show that LOHO is able to edit individual hair attributes, such as appearance and style (Figure~\ref{fig:supp_appearance_and_style}), and shape (Figure~\ref{fig:supp_shape}), while keeping other attributes unchanged.
LOHO is also able to manipulate multiple hair attributes jointly (Figure~\ref{fig:supp_mult_attr_1},\ref{fig:supp_mult_attr_2},\ref{fig:supp_mult_attr_3}).

\end{document}